\documentclass[10pt,twocolumn,letterpaper]{article}

\usepackage{cvpr}              
\newcommand\blfootnote[1]{
    \begingroup
    \renewcommand\thefootnote{}\footnote{#1}
    \addtocounter{footnote}{-1}
    \endgroup
}
\usepackage{multirow}
\definecolor{cvprblue}{rgb}{0.21,0.49,0.74}
\usepackage[pagebackref,breaklinks,colorlinks,allcolors=cvprblue]{hyperref}

\def\paperID{12608} 
\def\confName{CVPR}
\def\confYear{2025}

\title{Visual Persona: Foundation Model for Full-Body Human Customization}

\begin{document}
\author {
    Jisu Nam\textsuperscript{\rm 1}$^{*}$,
    Soowon Son\textsuperscript{\rm 1},
    Zhan Xu\textsuperscript{\rm 2},
    Jing Shi\textsuperscript{\rm 2},
    Difan Liu\textsuperscript{\rm 2}, 
    Feng Liu\textsuperscript{\rm 2}, 
    Aashish Misraa\textsuperscript{\rm 3}, \\
    Seungryong Kim\textsuperscript{\rm 1}$^\dag$,
    Yang Zhou\textsuperscript{\rm 2}$^\dag$ \\ \\
    \textsuperscript{\rm 1}KAIST AI \hspace{5pt}
    \textsuperscript{\rm 2}Adobe Research \hspace{5pt}
    \textsuperscript{\rm 3}Adobe \\
    {\tt\small\ \href{https:/cvlab-kaist.github.io/Visual-Persona}{https://cvlab-kaist.github.io/Visual-Persona}}
 \\
}

\newcommand{\paragrapht}[1]{\noindent\textbf{#1}}

\twocolumn[{
\renewcommand\twocolumn[1][]{#1}%
\maketitle
\thispagestyle{empty}
\vspace{-20pt}
\begin{center}
\includegraphics[width=0.95\textwidth]{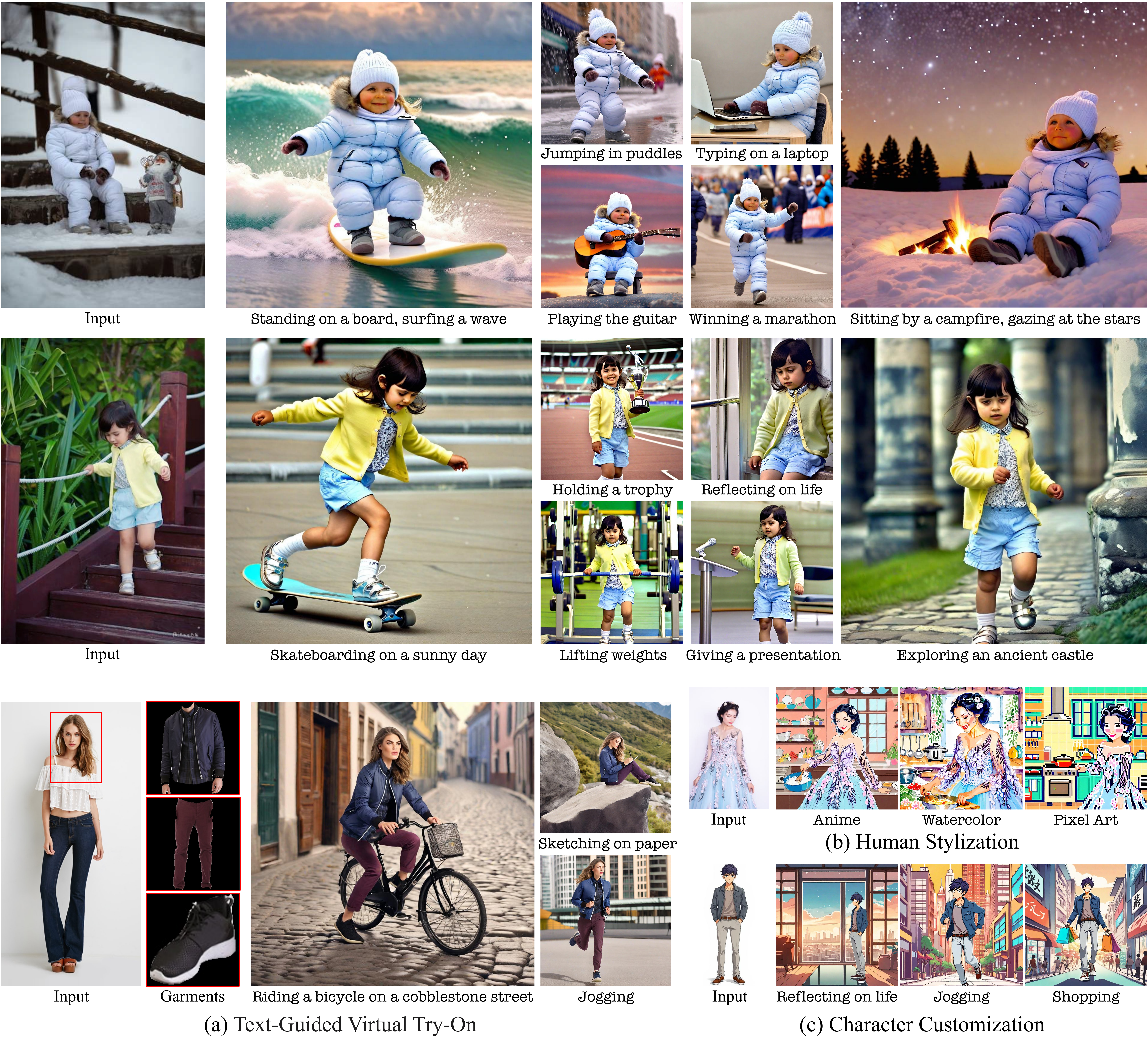}
\vspace{-3pt}
\captionof{figure}{\textbf{Visual Persona enables Text-to-Image Full-Body Human Customization.} Given a single human image, our method generates diverse customized images of the individual that are closely aligned with text descriptions while accurately preserving full-body appearance, which unlocks a wide range of applications, including (a) text-guided virtual try-on, (b) human stylization, and (c) character customization.}
\label{qual:teaser}
\end{center}
}]
\blfootnote{$^\dag$Co-corresponding author.}
\blfootnote{$^*$Work done during an internship at Adobe Research.}

\maketitle

\begin{abstract}
We introduce \emph{Visual Persona}, a foundation model for text-to-image full-body human customization that, given a single in-the-wild human image, generates diverse images of the individual guided by text descriptions. Unlike prior methods that focus solely on preserving facial identity, our approach captures detailed full-body appearance, aligning with text descriptions for body structure and scene variations. Training this model requires large-scale paired human data, consisting of multiple images per individual with consistent full-body identities, which is notoriously difficult to obtain. To address this, we propose a data curation pipeline leveraging vision-language models to evaluate full-body appearance consistency, resulting in \emph{Visual Persona-500K}—a dataset of 580k paired human images across 100k unique identities. For precise appearance transfer, we introduce a transformer encoder-decoder architecture adapted to a pre-trained text-to-image diffusion model, which augments the input image into distinct body regions, encodes these regions as local appearance features, and projects them into dense identity embeddings independently to condition the diffusion model for synthesizing customized images. \emph{Visual Persona} consistently surpasses existing approaches, generating high-quality, customized images from in-the-wild inputs. Extensive ablation studies validate design choices, and we demonstrate the versatility of \emph{Visual Persona} across various downstream tasks. 

\end{abstract}
    
\section{Introduction}
\label{sec:intro}

Recent advances in large-scale generative models~\cite{ho2020denoising, song2020denoising, saharia2022photorealistic, ramesh2022hierarchical, podell2023sdxl} have made remarkable strides in creating photorealistic images. The success of these models enables users to customize pre-trained models with their own personal data~\cite{gal2022image,ruiz2023dreambooth,ye2023ip,nam2024dreammatcher}. More recently, such methods have focused on human data~\cite{ye2023ip, xiao2023fastcomposer, gal2023encoder, wang2024instantid, guo2024pulid, peng2024portraitbooth, shi2024instantbooth, valevski2023face0, li2024photomaker, gal2024lcm, papantoniou2024arc2face, zhou2024storymaker}, driven by various practical applications, including film production~\cite{long2024videodrafter, jin2024appearance}, book illustration~\cite{tewel2024training, akdemir2024oracle}, and virtual/augmented reality~\cite{li2024aniartavatar, chen2024morphable, kim2024moditalker, nam2023diffusion}. These methods allow users to synthesize novel renditions of specific individuals.

However, most human-customized models primarily focus on generating human face images~\cite{ye2023ip, xiao2023fastcomposer, gal2023encoder, wang2024instantid, guo2024pulid, li2024photomaker, gal2024lcm, papantoniou2024arc2face}, restricting their applicability to the face domain. This limitation motivates us to broaden the scope of previous studies to the full-body human domain, which we refer to as \textit{Full-Body Human Customization}, a largely underexplored area. In this paper, we aim to develop a foundational model for full-body human customization, unlocking a wide range of in-the-wild applications.

We argue that such a foundational model should satisfy two key criteria: text alignment and identity preservation. Text alignment refers to the ability to generate diverse images aligned with given text descriptions, adjusting to variations in facial expressions, poses, actions, and surroundings. Identity preservation ensures that the model produces a consistent full-body appearance that accurately matches the input human image, including facial identity, clothing, and accessories. We present the development of this model, which embodies the above criteria, dubbed \textbf{Visual Persona}.

\begin{table}[b]
\captionsetup{font=small}
\small
\vspace*{-0.1in}
\begin{center}

\resizebox{3.3in}{!}{
    \setlength{\tabcolsep}{5pt}
    \renewcommand{\arraystretch}{1.1}
    \begin{tabular}{lcrccc}
    \toprule
    \textbf{Methods} & \textbf{Type} & \textbf{Database} & \textbf{Size} & \textbf{Res.} & \textbf{Domain}  \\
    \midrule
    IP-Adapter~\cite{ye2023ip} & Single & LAION-2B, COYO-700M  & 10M & - & General\\
    E4T~\cite{gal2023encoder} & Single & FFHQ, CelebA-HQ & 100K  & 1024 & Face\\
    PortraitBooth~\cite{peng2024portraitbooth} & Single & CelebV-T & 70K & $512 +$ & Face\\
    InstantID~\cite{wang2024instantid} & Single & LAION-Face, Internal & 60M & - & Face\\
    PuLID~\cite{guo2024pulid} & Single & Internal & 1.5M & - & Face\\
    InstantBooth~\cite{shi2024instantbooth} & Single & Internal  & 1.4M & $1024 +$ & Human \\
    StoryMaker~\cite{zhou2024storymaker} & Single & Internal  & 500K & - & Human \\
    \midrule
    LCM-Look.~\cite{gal2024lcm} & Paired & Synthetic & 500K(100K) & 1024 & Face\\
    PhotoMaker~\cite{li2024photomaker} & Paired & VoxCeleb1, VGGFace2 & 112K(13K) & $256 +$  & Face\\
    Arc2Face~\cite{papantoniou2024arc2face} & Paired & WebFace42M & 21M(1M)  & 448  & Face \\
    \midrule
    Visual Persona (Ours) & Paired & Visual Persona-500K & 580K(100K) & $1024 +$ & Human \\
    \bottomrule
    \end{tabular}
}
\vspace{-0.1in}
\caption{\textbf{Comparison of datasets in state-of-the-art customized models.} This table outlines the data type, database, data size, image resolution, and data domain used for each method. Unlike existing works that focus on the face domain or use a single image per individual, our approach aims to explore large-scale, real paired human data with full-body appearance consistency.}

\label{table:comparison}
\vspace*{-0.1in}

\end{center}
\end{table}

Large-scale paired human data, comprising multiple images of the same individuals with consistent full-body identities, is crucial for preserving full-body appearance consistency and achieving text alignment. However, obtaining such data is notoriously difficult. As a result, recent works use single human datasets that contain only one image per individual~\cite{shi2024instantbooth, zhou2024storymaker}. To overcome this, we design a data curation pipeline that constructs paired human data from a large pool of unpaired human collections. In light of recent progress in Vision Language Models (VLMs)~\cite{wang2023cogvlm, achiam2023gpt, bai2023qwen, lu2024deepseek, liu2024visual, team2023gemini, abdin2024phi}, we propose that VLMs can serve as simple yet effective tools for evaluating full-body visual consistency, even when considerable variations exist across images. Furthermore, we generate detailed text captions for each image to disentangle the individual’s identity from intra-individual variations depicted in the text prompts. As a result, we collected a total of 580k images with text captions across 100k unique full-body identities, which we denote as \textbf{Visual Persona-500K}. In Table~\ref{table:comparison}, we highlight that Visual Persona-500K surpasses the scope of earlier datasets.

For identity preservation, it is important to accurately transfer the input full-body human's appearance to the customized image while retaining the large geometric deformations introduced by pre-trained Text-to-Image (T2I) models. Recent works~\cite{gal2023encoder, ye2023ip, xiao2023fastcomposer, valevski2023face0, gal2024lcm, wang2024instantid, guo2024pulid, shi2024instantbooth, li2024photomaker, papantoniou2024arc2face, peng2024portraitbooth, zhou2024storymaker} solved human customization using pre-trained image encoders to accelerate the tuning process. These approaches commonly leverage semantic representations from CLIP~\cite{radford2021learning} or facial recognition models~\cite{deng2020retinaface, deng2019arcface, serengil2020lightface, serengil2024lightface} to encode the input image, then globally map the encoded features to compact identity embeddings through simple linear projection layers~\cite{ye2023ip, li2024photomaker, gal2023encoder, xiao2023fastcomposer, guo2024pulid, peng2024portraitbooth}. While effective in the face domain, they often fail to preserve the detailed appearance of in-the-wild human inputs.

In this paper, we propose a transformer encoder-decoder architecture, adapted to a pre-trained T2I diffusion model, specifically designed for full-body human customization. To effectively transfer each body part of the input human into the complex body structure of the customized images, we augment the input image into distinct body images using an off-the-shelf parsing method~\cite{li2020self}. A pre-trained image transformer encoder~\cite{oquab2023dinov2} encodes the set of body images for the individual and extracts local appearance features from each. The body-partitioned transformer decoder then projects each body feature into its corresponding identity embedding through a cross-attention layer, further refining intra-relationships within the embeddings using a self-attention layer. The output embeddings are concatenated along the token length to condition the pre-trained T2I diffusion model through a learnable identity cross-attention module, while text embeddings condition the model via a text cross-attention module. Finally, the diffusion model synthesizes diverse images with a consistent identity matching the input. Notably, we freeze the pre-trained T2I model to maximize its generative capability, preserving both image quality and text alignment.

We evaluate the effectiveness of our method on a GPT-based evaluation~\cite{peng2024dreambench++} and a human evaluation, with comparison to state-of-the-art works~\cite{ye2023ip, wang2024instantid, li2024photomaker, zhou2024storymaker}. Comprehensive ablation studies validate our design choices and the contribution of each component. Furthermore, we highlight the versatility of our method across various applications.

\section{Related Work}
\label{sec:related_works}

\paragrapht{Learning Domain Priors for T2I Customization.} The objective of T2I customization~\cite{gal2022image, ruiz2023dreambooth, kumari2023multi} is to adapt pre-trained T2I generative models to user-provided subject images. While earlier tuning-based approaches~\cite{gal2022image, ruiz2023dreambooth, dong2022dreamartist, alaluf2023neural, voynov2023p+, kumari2023multi, han2023svdiff, ryu2023lora} involve a laborious tuning process for each subject, recent methods have shifted toward learning domain-specific priors from large-scale data within a single domain~\cite{wei2023elite, gal2023encoder, ye2023ip, xiao2023fastcomposer, valevski2023face0, chen2024anydoor, gal2024lcm, wang2024instantid, guo2024pulid, shi2024instantbooth, li2024photomaker, peng2024portraitbooth, zhou2024storymaker, kim2025diffface}. In particular, human face customization has gained significant attention in this area~\cite{gal2023encoder, ye2023ip, xiao2023fastcomposer, valevski2023face0, gal2024lcm, wang2024instantid, guo2024pulid, shi2024instantbooth, li2024photomaker, peng2024portraitbooth, papantoniou2024arc2face}, driven by extensive face datasets, such as FFHQ~\cite{karras2019style}, CelebA-HQ~\cite{karras2017progressive}, CelebV-T~\cite{yu2023celebv}, and LAION-Face~\cite{schuhmann2021laion}. However, these datasets are confined to the face domain, leaving the full-body human domain mostly unexplored. Although some works~\cite{shi2024instantbooth, zhou2024storymaker} seek to use in-the-wild human images, it is challenging to obtain paired human data consisting of multiple images per individual with consistent full-body identity. As a result, these methods compromise by training on a single image per individual, which struggles with the trade-off between identity preservation and text alignment~\cite{guo2024pulid, gal2024lcm, papantoniou2024arc2face}.

\paragrapht{Encoder-based T2I Customization.} To initialize the tuning process, domain-tuning approaches for face customization~\cite{gal2023encoder, ye2023ip, xiao2023fastcomposer, valevski2023face0, gal2024lcm, wang2024instantid, guo2024pulid, shi2024instantbooth, li2024photomaker, papantoniou2024arc2face, peng2024portraitbooth} often extract semantic representations from pre-trained image encoders such as CLIP~\cite{radford2021learning} or face recognition models~\cite{szegedy2015going, szegedy2017inception, huang2020curricularface, deng2019arcface}, which are trained with global supervision on text or identity classes. These methods commonly embed features from the image encoder into a small number of token embeddings~\cite{gal2023encoder, valevski2023face0, shi2024instantbooth, xiao2023fastcomposer, li2024photomaker, peng2024portraitbooth, papantoniou2024arc2face, ye2023ip, guo2024pulid, wang2024instantid, gal2024lcm}, often using linear projection layers~\cite{ye2023ip, li2024photomaker, gal2023encoder, xiao2023fastcomposer, guo2024pulid, peng2024portraitbooth}. Although effective in the face domain, these approaches often struggle with full-body human inputs, as globally mapping into a small number of tokens fails to condense key visual information and tends to blend local details from each body part, resulting in poor full-body appearance preservation.

\begin{figure*}[t]
    \begin{center}
\includegraphics[width=1.0\textwidth]{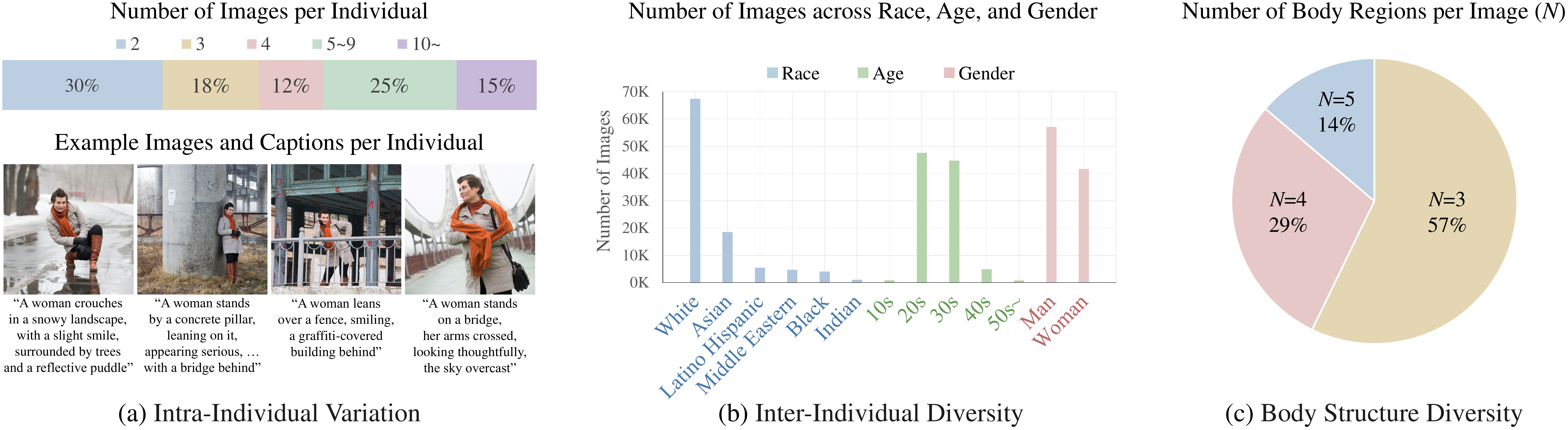}
    \end{center}
    \vspace{-5pt}
    \caption{\textbf{Data Statistics:} Our curated training dataset, \textbf{Visual Persona-500K}, consists of 580k images representing 100k individuals. (a) illustrates the distribution of the number of images per individual, with over 50\% of individuals having more than four images, and shows example image-caption pairs from the same individual. (b) highlights the diversity of individuals based on facial attributes, including race, age, and gender, which are estimated by DeepFace~\cite{serengil2021lightface}. (c) showcases body structure diversity, segmented into five clusters—full-body, face, torso, legs, and shoes—categorized using a body-parsing method~\cite{li2020self}.}
    
    \label{qual:data_statistics}
    \vspace{-10pt}
\end{figure*}

\section{Method}
\label{sec:method}

\subsection{Visual Persona-500K Dataset}
\label{sec:dataset} 
A paired human dataset, comprising multiple images per individual with consistent identities, is essential for concurrently achieving identity preservation and text alignment. However, due to the challenges of assembling consistent full-body human pairs, previous works have compromised by training on a single image per individual~\cite{shi2024instantbooth, zhou2024storymaker}. In this work, we curated our paired human dataset, Visual Persona-500K, from a large pool of unpaired human collections that include visually inconsistent images of the same individuals. Furthermore, we generate detailed captions for each image to disentangle an individual’s identity from input and intra-individual variations driven by text prompts. The overall statistics and scope of our dataset are presented in Figure~\ref{qual:data_statistics} and Table~\ref{table:comparison}. 

\paragrapht{Curating Consistent Facial Identities.} We first collect unpaired human data, consisting of multiple images per each individual. To further ensure facial identity consistency, we calculate the cosine similarity between facial embeddings for each image pair using a face recognition model~\cite{deng2019arcface}. We then select an anchor image with the highest average similarity score and discard any images whose similarity scores with the anchor image fall below a predefined threshold. We further filter out images with unidentifiable faces, watermarks, or duplicates. For high image quality, we discard images where the shortest side is less than 1024 pixels. 

\paragrapht{Curating Consistent Full-Body Identities.} From the unpaired human data, which only guarantees facial identity consistency, we further evaluate body identity consistency by assessing whether each individual is wearing identical clothing. Specifically, we suggest that VLMs~\cite{liu2024visual, abdin2024phi, bai2023qwen, wang2023cogvlm, lu2024deepseek, team2023gemini, achiam2023gpt} are simple yet powerful tools for evaluating the visual consistency of the human body. For each image subset of the same individual, we prompt LLAVA~\cite{liu2024visual} to assess whether the individuals in the subset are wearing the same clothing. A simple prompt—\textit{“Are they wearing exactly the same clothes?”}—triggers the model to provide a binary decision with high precision. If the model returns a positive response for all subsets of the individual, the individual is retained; otherwise, the individual is excluded from the dataset. In total, we curated a dataset of 580k full-body paired human images across 100k unique individuals, referred to as Visual Persona-500K. A detailed description of the curation method is provided in Appendix~\ref{sup:VP-500K}.

\paragrapht{Captioning.} 
We aim to disentangle the individual's identity in the input from intra-individual variations depicted in the text prompts. To achieve this, we use Phi-3~\cite{abdin2024phi} to generate detailed text descriptions for each image that contain identity-irrelevant information. The prompt used is: \textit{“Describe the image in detail in one sentence, focusing on facial expression, pose, actions, and surroundings.”} Notably, our data curation and captioning pipeline is compatible with any off-the-shelf VLM~\cite{liu2024visual, abdin2024phi, bai2023qwen, wang2023cogvlm, lu2024deepseek, team2023gemini, achiam2023gpt}.

\begin{figure*}[htbp]
    \begin{center}
\includegraphics[width=0.99\textwidth]{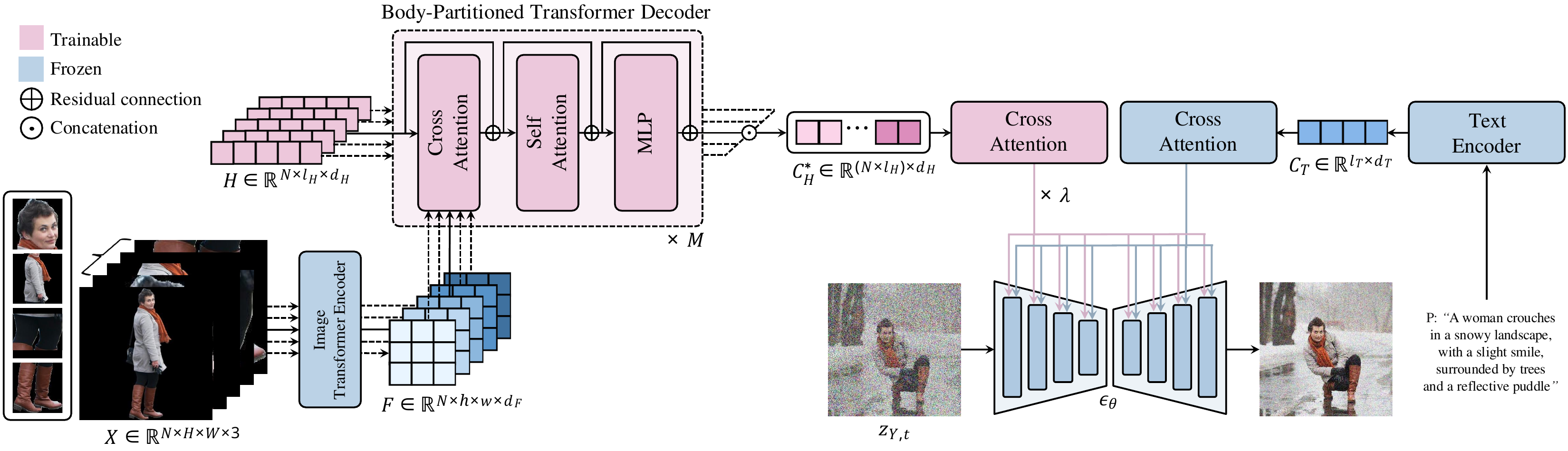} 
    \end{center}
    \vspace{-15pt}
    \caption{\textbf{Overall Architecture:} Our network augments the input human image into body regions, which are encoded into local features by an image transformer encoder. A body-partitioned transformer decoder projects these features into learnable identity embeddings via cross-attention, followed by self-attention and MLP. After $M$ iterations, the embeddings are concatenated to form a stacked identity embedding. The identity embedding and text embedding from detailed captions condition a pre-trained T2I diffusion model to synthesize a new image with the input identity. Only the body-partitioned transformer decoder and identity cross-attention module are trained.}
    \vspace{-10pt}
    \label{qual:overall}
\end{figure*}

\begin{figure}[t]
    \centering
    \includegraphics[width=0.45\textwidth]{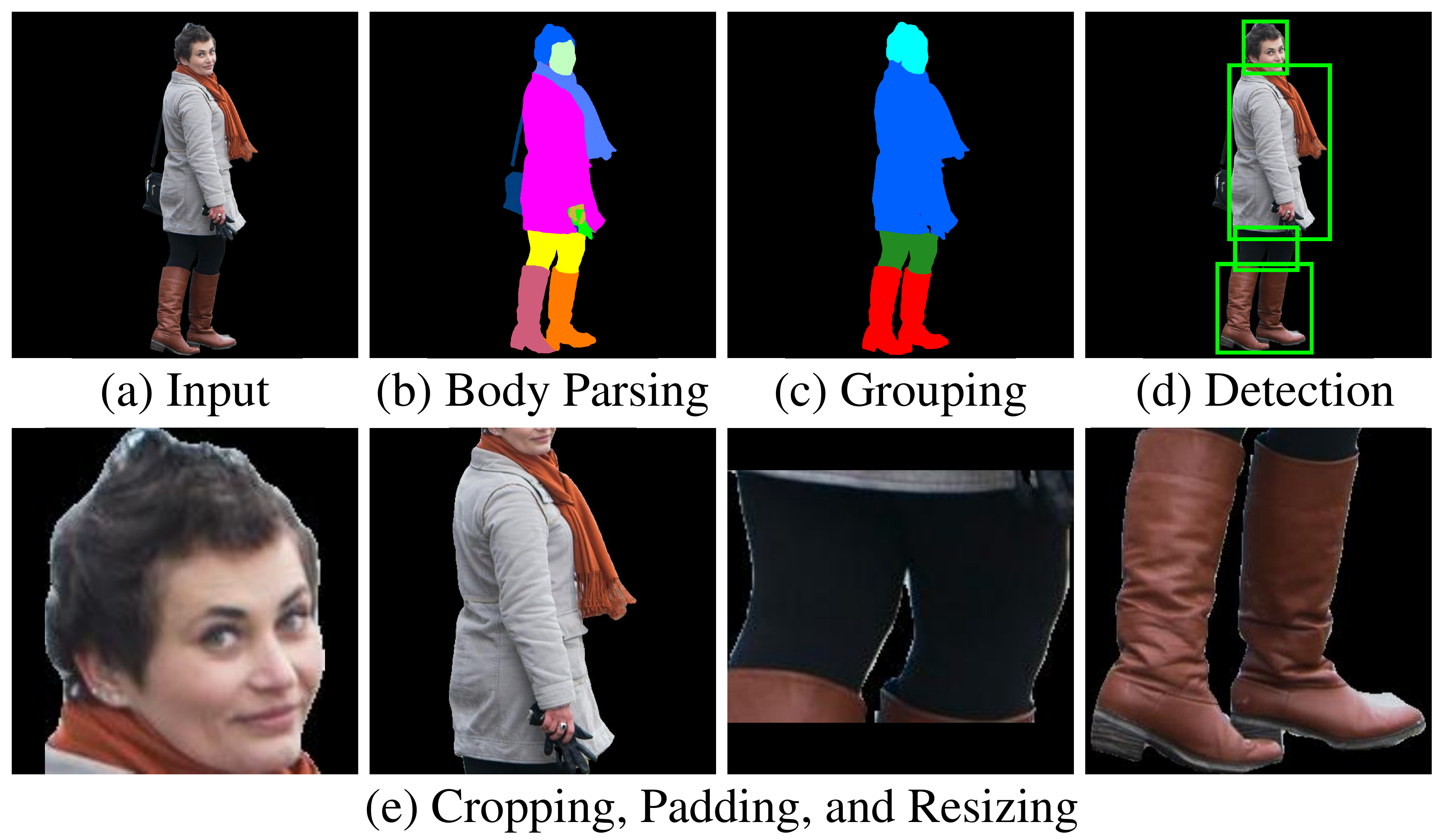} 
    \caption{\textbf{Body Part Decomposition.}}
    
    \label{qual:body_part}
\end{figure}

\subsection{Model Architecture}
\label{sec:architecture} 
Given a single human image and a text prompt, our goal is to transfer the full-body appearance from the input to customized images generated by the pre-trained T2I model, while preserving its generative capabilities for image quality and text alignment. Prior studies~\cite{gal2023encoder, ye2023ip, xiao2023fastcomposer, valevski2023face0, gal2024lcm, wang2024instantid, guo2024pulid, shi2024instantbooth, li2024photomaker, papantoniou2024arc2face, peng2024portraitbooth, zhou2024storymaker} commonly leverage semantic representations from pre-trained image encoders, such as CLIP~\cite{radford2021learning}, by mapping them into compact identity embeddings, often through linear projection layers~\cite{ye2023ip, li2024photomaker, gal2023encoder, xiao2023fastcomposer, guo2024pulid, peng2024portraitbooth}. However, these methods frequently struggle to preserve the detailed appearance of in-the-wild human inputs.

In this paper, we introduce a novel transformer encoder-decoder architecture~\cite{vaswani2017attention}, adapted to a pre-trained T2I diffusion model, specifically designed for full-body human customization. We decompose the input human body into distinct body regions as separate images. The image transformer encoder then encodes each body image into localized appearance features, while the body-partitioned transformer decoder projects each feature into the corresponding dense identity embeddings. These embeddings then guide the pre-trained T2I model to synthesize diverse customized images that accurately preserve the input’s full-body appearance. The overall architecture is illustrated in Figure~\ref{qual:overall}. 

\paragrapht{Body Part Decomposition.} 
Figure~\ref{qual:body_part} provides an overview of body part decomposition. To effectively transfer the input human appearance to the complicated body structure in the customized image, it is important for the diffusion model to attend independently to each distinct body part from the input and map it to the corresponding part in the synthesized image. To achieve this, we augment the input image $I \in \mathbb{R}^{H \times W \times 3}$ (Figure~\ref{qual:body_part}(a)), where $H$, $W$, and 3 represent the height, width, and RGB channels, respectively, into $N$ distinct body images. Here, we apply a foreground mask~\cite{huynh2024maggie} to the input image to focus solely on the human part. Specifically, we leverage an off-the-shelf body parsing method~\cite{li2020self} to parse body regions (Figure~\ref{qual:body_part}(b)) and group them into $N-1$ categories (Figure~\ref{qual:body_part}(c)). We then extract bounding box coordinates for each category (Figure~\ref{qual:body_part}(d)), crop the corresponding regions from $I$, zero-pad along the largest side to preserve image scale, and resize them to the original input size (Figure~\ref{qual:body_part}(e)). This results in an input set of the individual, $X \in \mathbb{R}^{N \times H \times W \times 3}$, comprising one full-body image and $N-1$ body part images. Note that $N$ can represent any number of body regions. In this paper, we empirically set $N$ to 5, corresponding to the full body, face, torso, legs, and shoes. Table~\ref{tab:comp_analysis} demonstrates the effectiveness of body part decomposition.

\paragrapht{Image Transformer Encoder.} For detailed appearance preservation, we extract local image features from the input set of the individual. Specifically, we use the pre-trained vision transformer DINOv2~\cite{oquab2023dinov2} as our image encoder. In contrast to other image encoders~\cite{radford2021learning, he2016deep} trained with weak supervision on image-text or class alignment, which focus on the semantics of images, the self-supervised vision transformer DINOv2~\cite{oquab2023dinov2} can capture fine-grained, localized features, including structural and texture information across images~\cite{jiang2023clip, zhang2024tale}. This makes DINOv2~\cite{oquab2023dinov2} particularly well-suited for our task of preserving visual details, such as clothing patterns and textures. Specifically, we encode the input set $X$ into a set of image features $F \in \mathbb{R}^{N \times h \times w \times d_F}$, where $h$, $w$, and $d_F$ represent the height, width, and channel dimensions of each feature. Notably, instead of using the pre-defined class token {\ttfamily CLS}, we leverage the full set of local tokens to focus on spatial information from the input.

\paragrapht{Body-Partitioned Transformer Decoder.} A straightforward approach to mapping the input features onto the pre-trained T2I model involves using Multi-Layer Perceptron (MLP) layers, as in prior works~\cite{ye2023ip, li2024photomaker, gal2023encoder, xiao2023fastcomposer, guo2024pulid, peng2024portraitbooth}. However, as discussed in Table~\ref{tab:comp_analysis}, simply using an MLP is insufficient to capture key visual attributes from complex input features, particularly from in-the-wild human images, and tends to blend local details across the body into a global appearance. To address this, we propose a body-partitioned transformer decoder that projects local features from each body region into corresponding dense identity embeddings, preserving detailed visual appearance. Each transformer layer comprises a cross-attention layer, a self-attention layer, and an MLP layer.

Let $H \in \mathbb{R}^{N \times l_{H} \times d_{H}}$ represent the learnable hidden identity embeddings for each body image, whose dimensions match those of the output identity embeddings. Here, $l_{H}$ and $d_{H}$ denote the token length and the channel dimension, respectively. We highlight that, unlike prior methods~\cite{gal2023encoder, ye2023ip, xiao2023fastcomposer, valevski2023face0, gal2024lcm, wang2024instantid, guo2024pulid, shi2024instantbooth, li2024photomaker, papantoniou2024arc2face, peng2024portraitbooth, zhou2024storymaker} that map image features into a small number of tokens (typically $l_{H}=16$), we set $l_{H} = h \times w$, where $h$ and $w$ represent the height and width of the DINOv2 feature. Table~\ref{quan:token_length} shows that dense embeddings preserve finer full-body details that a small number of tokens may overlook. For the $i$-th input body image and the $j$-th transformer layer, the cross-attention layer links each hidden embedding $H^{i,j}$ to its corresponding feature $F^{i}$ from the $i$-th body image $X^{i}$. The updated embedding, $H^{i,j}_{{ca}}$, is then passed through the self-attention layer, which learns internal relationships within the embedding. The resulting embedding, $H^{i,j}_{{sa}}$, is further processed by the MLP layer, mapping it to the output embedding $H^{i,j+1}$, which is recursively fed into the next transformer layer. This process is formulated as:
\begin{equation}
H^{i,j}_{{ca}}= \mathtt{C\texttt{-}Att}(\mathtt{LN}(H^{i,j}), F^i, F^i) + H^{i,j}, 
\end{equation}
\begin{equation}
H^{i,j}_{{sa}} = \mathtt{S\texttt{-}Att}(\mathtt{LN}(H^{i,j}_{{ca}}), \mathtt{LN}(H^{i,j}_{{ca}}), \mathtt{LN}(H^{i,j}_{{ca}})) + H^{i,j}_{{ca}}, \end{equation}
\begin{equation}
H^{i,j+1} = \mathtt{MLP}(\mathtt{LN}(H^{i,j}_{{sa}})) + H^{i,j}_{{sa}},
\end{equation}
where $\mathtt{LN}$ stands for layer normalization. $\mathtt{C\texttt{-}Att}(\cdot)$ represents the cross-attention layer, where $\mathtt{LN}(H^{i,j})$ is projected as the queries and $F^i$ is projected as the keys and values. $\mathtt{S\texttt{-}Att}(\cdot)$ denotes the self-attention layer, where $\mathtt{LN}(H^{i,j}_{{ca}})$ is projected as the queries, keys, and values. $\mathtt{MLP}$ refers to the multi-layer perceptron layer. After $M$ iterations, we obtain the refined identity embeddings, $C_H \in \mathbb{R}^{N \times l_{H} \times d_{H}}$. Finally, these embeddings are concatenated along the token length to form a stacked identity embedding, $C^*_H$:
\begin{align}
C^*_H = \mathtt{Concat}([C^1_H, \dots, C^N_H]) \in \mathbb{R}^{(N \times l_{H}) \times d_H},
\end{align}
where $\mathtt{Concat}(\cdot)$ represents the concatenation operation.

\paragrapht{Decoupled Cross-Attention.} Given $C^*_H$, the pre-trained T2I model $\epsilon_{\theta}(\cdot)$ portrays the target image $Y$, aligned with the text embedding $C_T$ derived from text prompt $T$. Specifically, we adopt a decoupled cross-attention mechanism~\cite{ye2023ip}. $C^*_H$ is processed through an additional learnable identity cross-attention module, where target queries $Q_Y$ for $Y$ attend to keys $K^*_H$ and values $V^*_H$ projected from $C^*_H$. The text embedding $C_T \in \mathbb{R}^{l_{T} \times d_{T}}$ is incorporated through the text cross-attention module, projected as keys $K_T$ and values $V_T$, where $l_{T}$ and $d_{T}$ denote the token length and the channel dimension, respectively. The output $Z'$ is formulated as:
\begin{align} 
\label{attention_ip}
{Z'} = \mathtt{C\texttt{-}Att}({Q_Y}, {K_T}, {V_T}) + \lambda \cdot \mathtt{C\texttt{-}Att}({Q_Y}, {K^*_H}, {V^*_H}), 
\end{align}
where $\mathtt{C\texttt{-}Att}(\cdot)$ is the text or identity cross-attention module, and $\lambda$ is the weighting scalar.

\subsection{Training}
\label{sec:training} 
We solely train the body-partitioned transformer decoder and the identity cross-attention module, while freezing all other parameters, including the image transformer encoder, body parsing model, and T2I diffusion model. For training, we leverage our paired human dataset, Visual Persona-500K, where the input set of the individual, $X$, guides the frozen diffusion model $\epsilon_{\theta}(\cdot)$ to render another image of the same individual, $Y$, aligned with the target prompt $T$. The training objective is formulated by minimizing the prediction error ${L}$~\cite{ho2020denoising, song2020denoising}:
\begin{equation} 
\label{equ:training}
{L} := \mathbb{E}_{z_{Y,t},\epsilon,t,{C}_T,C^*_H} \left[ \left\lVert \epsilon - \epsilon_{\theta} (z_{Y,t}, t, {C}_T, C^*_H) \right\rVert^2_2 \right],
\end{equation}
where $\epsilon$ is the added noise in the forward pass and $z_{Y,t}$ represents the noisy latent variable of $Y$ at time step $t$. ${L}$ is backpropagated through the diffusion model, updating the parameters in the body-partitioned transformer decoder and the identity cross-attention module. Note that prior works~\cite{gal2023encoder, ye2023ip, xiao2023fastcomposer, valevski2023face0, wang2024instantid, guo2024pulid, shi2024instantbooth, peng2024portraitbooth, zhou2024storymaker} were trained on identical images for $X$ and $Y$, leading to overfitting on identity-unrelated elements like background, lighting, and composition~\cite{guo2024pulid, gal2024lcm, papantoniou2024arc2face}.

\begin{figure*}[t]
    \begin{center}
\includegraphics[width=1.0\textwidth]{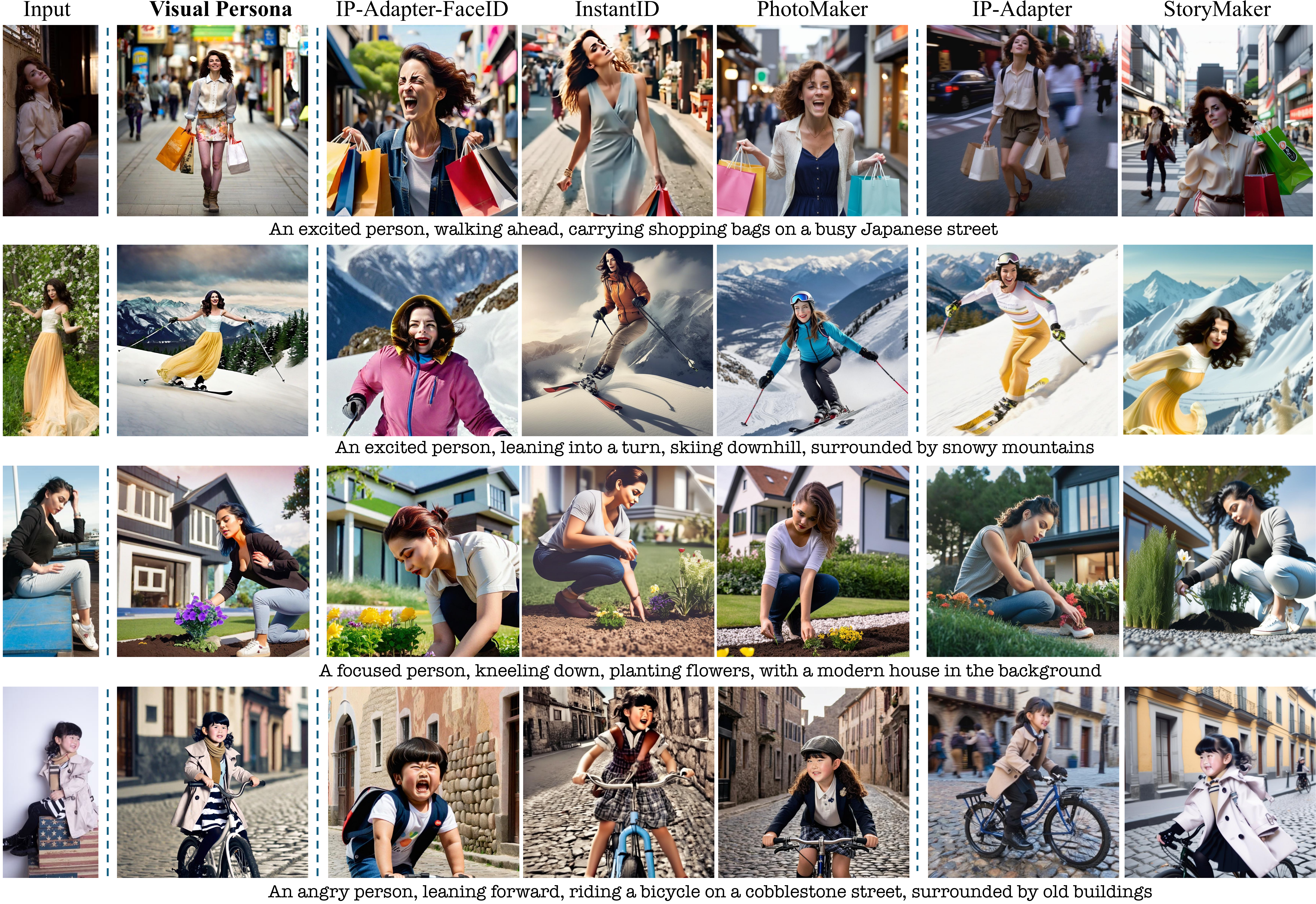} 
    \end{center}
    \vspace{-15pt}
    \caption{\textbf{Qualitative Comparison on PPR10K~\cite{liang2021ppr10k}:} Compared to prior works that focus on face identity preservation~\cite{ye2023ip, wang2024instantid, li2024photomaker} or fail to capture the input's detailed appearance~\cite{ye2023ip, zhou2024storymaker}, Visual Persona accurately preserves the full-body appearance while generating diverse images based on text prompts.}
    \vspace{-10pt}
    \label{qual:main_qual}
\end{figure*}

\section{Experiments}
\label{sec:experiments}
\subsection{Experimental Settings}

\paragrapht{Dataset.} We evaluated our method on two human datasets: SSHQ~\cite{fu2022stylegan} and PPR10K~\cite{liang2021ppr10k}. SSHQ~\cite{fu2022stylegan} includes high-quality full-body images featuring various identities, poses, and clothing, sourced from DeepFashion~\cite{liu2016deepfashion} and African images from InFashAI~\cite{hacheme2021neural}. PPR10K~\cite{liang2021ppr10k} consists of a wide range of in-the-wild human images with significant geometric variations and diverse viewpoints. We randomly selected 50 individuals each from the SSHQ and PPR10K test splits~\cite{shi2024instantbooth} for testing. To assess text alignment, we augmented 17 prompts for live objects in Dreambooth~\cite{ruiz2023dreambooth} using ChatGPT~\cite{achiam2023gpt} to include facial expressions, poses, actions, and surroundings. For evaluation, all methods generated 4 samples for each input image and prompt pair, resulting in 3,400 samples per evaluation dataset. Additional details on the evaluation data are provided in Appendix~\ref{supp:eval_dataset}.

\paragrapht{Evaluation Metric.} As discussed in~\cite{christodoulou2024finding, tan2024evalalign, li2024genai, jiang2024genai, lin2025evaluating} and Appendix~\ref{supp:metrics}, obtaining reliable quantitative evaluations for customized image generation is challenging, as they heavily rely on human preferences. In this paper, we evaluated our method using Dreambench++~\cite{peng2024dreambench++}, an automated, human-aligned, GPT~\cite{achiam2023gpt}-based benchmark.

To evaluate identity preservation, we adapted Dreambench++~\cite{peng2024dreambench++}, prompting GPT~\cite{achiam2023gpt} to assess face identity, clothing type, design, texture, and color in generated images compared to input images, denoted as D-I. To evaluate text alignment, we instructed GPT to assess pose, actions, surroundings, composition, and overall quality to ensure prompt alignment, denoted as D-T. Each metric assigns an integer score from 0 (no resemblance or correlation) to 9 (near-perfect resemblance or correlation). As we aim for high scores on both D-I and D-T, following~\cite{xian2018zero, bucher2019zero}, we also calculate their harmonic mean per sample, denoted as D-H. As GPT has difficulty detecting facial expressions when the subject is distant from the foreground, we also conducted a human evaluation for text alignment on facial expressions, detailed in Appendix~\ref{supp:metrics}.

\paragrapht{Human Evaluation Setting.} 
We conducted a rigorous human study, strictly following the ImagenHub evaluation protocol~\cite{ku2023imagenhub}. Each generated sample, along with its input image and text prompt, was assessed by two metrics: semantic consistency (SC) and perceptual quality (PQ), both scored as \{0, 0.5, 1\}. For SC, 0 indicates that either the text or the identity does not align with the generated sample, 0.5 indicates partial alignment of both, and 1 indicates full alignment. This ensures a balanced evaluation, penalizing bias toward either text alignment or identity preservation. PQ measures the visual realism of the generated image, where higher is better. The final score is calculated as $O = \sqrt{SC \times PQ}$. Following~\cite{hu2024instruct}, eight human raters were recruited and trained according to ImagenHub~\cite{ku2023imagenhub} guidelines. Eight raters were divided into two groups, with each group evaluating 150 samples generated by three methods using the same input images and prompts to ensure rating consistency. Further details are in Appendix~\ref{supp:human_evaluation}.

\begin{table}[t]
\centering
\resizebox{1.0\columnwidth}{!}{%
\begin{tabular}{l|ccc|ccc}
    \toprule
     \multirow{2}{*}{Method} & \multicolumn{3}{c|}{SSHQ} & \multicolumn{3}{c}{PPR10K} \\ 
     & \multicolumn{1}{c}{D-I $\uparrow$} & \multicolumn{1}{c}{D-T $\uparrow$} & \multicolumn{1}{c|}{D-H $\uparrow$} & \multicolumn{1}{c}{D-I $\uparrow$} & \multicolumn{1}{c}{D-T $\uparrow$} & \multicolumn{1}{c}{D-H $\uparrow$} \\
    \midrule
    
    IP-Adapter-FaceID~\cite{ye2023ip} &  1.78 & 7.50 & 2.76 &  1.86 & 7.49 & 2.81
    \\InstantID~\cite{wang2024instantid} & 1.52 & 6.94 & 2.37 & 1.70 & 7.12 & 2.63\\
    PhotoMaker~\cite{li2024photomaker} & 1.70 & \textbf{7.72} & 2.64 & 2.03 & \textbf{7.64} & 3.03 \\ 
    
    \midrule
    IP-Adapter~\cite{ye2023ip} &  5.44 & 7.26 & 5.96& 5.04 & 7.38 & 5.75 \\
    StoryMaker~\cite{zhou2024storymaker} & 6.74 & 7.08 & 6.71 &  6.80 & 6.77 & 6.63 \\
    \textbf{Visual Persona (Ours)} & \textbf{7.10} & 7.15 & \textbf{6.99} & \textbf{7.30} & 6.67 & \textbf{6.85} \\

    \bottomrule
\end{tabular}%
}
\vspace{-5pt}
\caption{\textbf{Quantitative Comparison.}}
\label{quan:gpt}
\end{table}
\begin{figure}[t]
    \centering
    \vspace{-5pt}
    \includegraphics[width=0.4\textwidth]{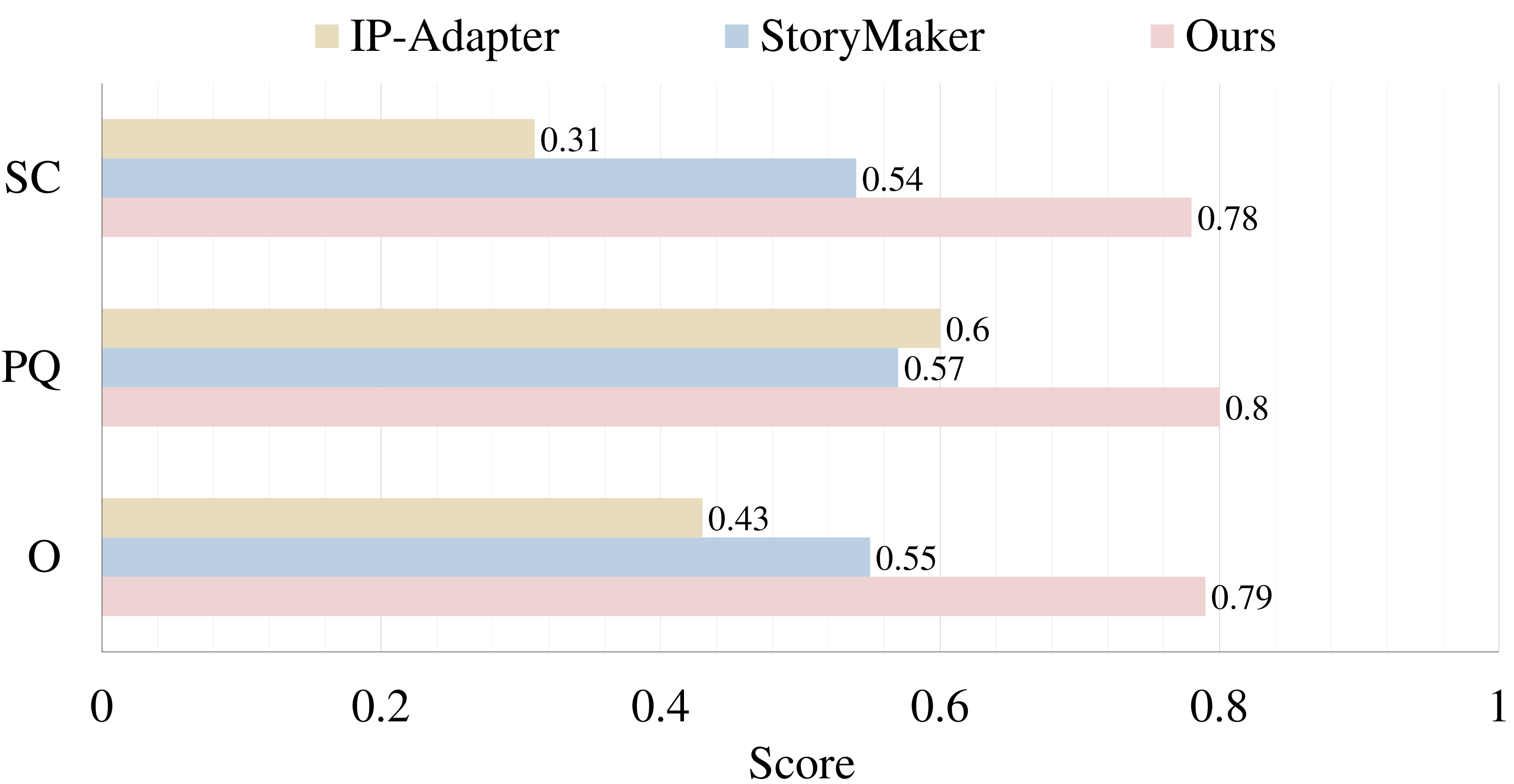} 
    \vspace{-5pt}
    \caption{\textbf{Human Evaluation.} }
    \vspace{-5pt}
    
    \label{qual:human}
\end{figure}

\begin{figure*}[t]
    \begin{center}
\includegraphics[width=1\textwidth]{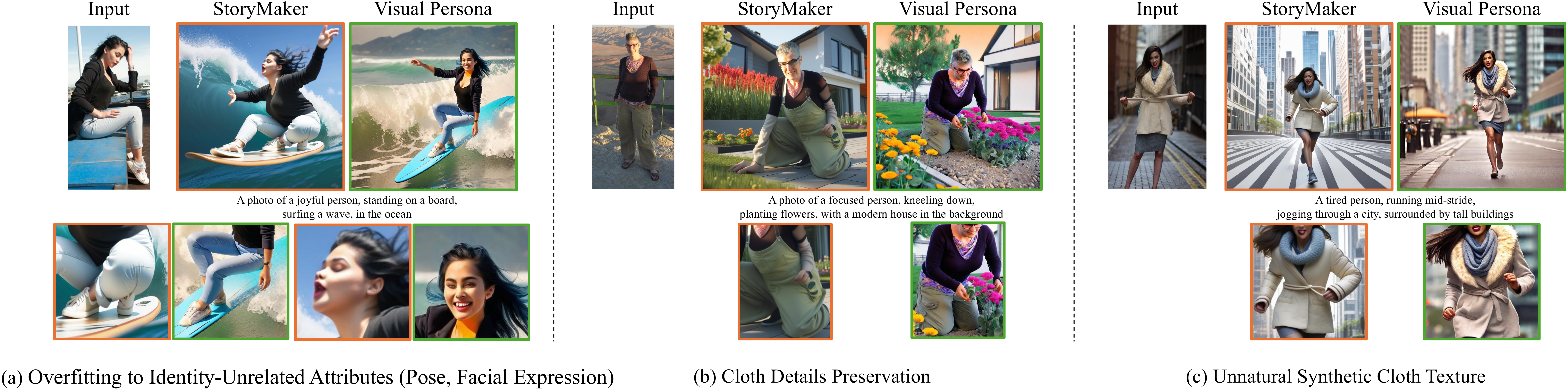} 
    \end{center}
    \vspace{-15pt}
    \caption{\textbf{Comparison between StoryMaker~\cite{zhou2024storymaker} ({\color{Orange} orange}) and Visual Persona ({\color{Green}green}), including full and zoomed-in images:} Compared to StoryMaker, Visual Persona enables large deformations, including pose and facial expression variations, preserves clothing details, and generates realistic clothing textures.}
    \label{qual:comparison_storymaker}
\end{figure*}

\begin{table*}[h]
\centering
\resizebox{0.9\linewidth}{!}
{
\begin{tabular}{l|cc|ccc}
    \toprule
     \multirow{2}{*}{Method} & \multicolumn{2}{c|}{\textbf{(a) Training}} & \multicolumn{3}{c}{\textbf{(b) Model Architecture}} \\ 
     & \multicolumn{1}{c}{Dataset} & \multicolumn{1}{c|}{Strategy} & \multicolumn{1}{c}{Body Part Decomposition} & \multicolumn{1}{c}{Encoder} & \multicolumn{1}{c}{Decoder} \\
    \midrule
    
    StoryMaker~\cite{zhou2024storymaker} & Unpaired & Reconstruction & 2 (Face, Body) & CLIP~\cite{radford2021learning}, buffalo\_l~\cite{deng2019arcface}
 & Resampler, Linear \\
    {Visual Persona} & Paired & Cross-Image & 5 (Full-Body, Face, Torso, Legs, Shoes) & DINO~\cite{oquab2023dinov2} & Transformer Decoder \\

    \bottomrule
\end{tabular}%
}
\vspace{-5pt}
\caption{\textbf{Comparison between StoryMaker~\cite{zhou2024storymaker} and Visual Persona.}}
\label{table:storymaker}
\end{table*}

\subsection{Results}
\paragrapht{Comparison.} Figure~\ref{qual:main_qual} and Table~\ref{quan:gpt} summarize the qualitative and quantitative comparisons with state-of-the-art customized models~\cite{ye2023ip, li2024photomaker, wang2024instantid, zhou2024storymaker}. As shown in Figure~\ref{qual:main_qual}, face customization models~\cite{ye2023ip, li2024photomaker, wang2024instantid} exhibit strong text alignment but are limited to the face domain, while full-body customization models~\cite{ye2023ip, zhou2024storymaker} often fail to preserve the detailed appearance of inputs and produce foreground-biased outputs. In contrast, Visual Persona accurately retains the input's full-body appearance while adapting to complex body deformations and scene changes driven by the text prompts. Table~\ref{quan:gpt} further shows that Visual Persona significantly outperforms previous methods in identity preservation (D-I), while maintaining comparable text alignment (D-T), ultimately achieving the best harmonic mean (D-H). More results are provided in Appendix~\ref{supp:more_results}.

\paragrapht{Human Evaluation.} We present the human evaluation results in Figure~\ref{qual:human}. Unlike previous methods that are biased toward text alignment~\cite{ye2023ip} or produce artificial outputs~\cite{zhou2024storymaker}, Visual Persona surpasses these approaches in SC, which concurrently measures identity preservation and text alignment, as well as in PQ, which evaluates image quality, achieving the highest overall score O.

\begin{figure}[t]
    \centering
    \includegraphics[width=1.0\linewidth]{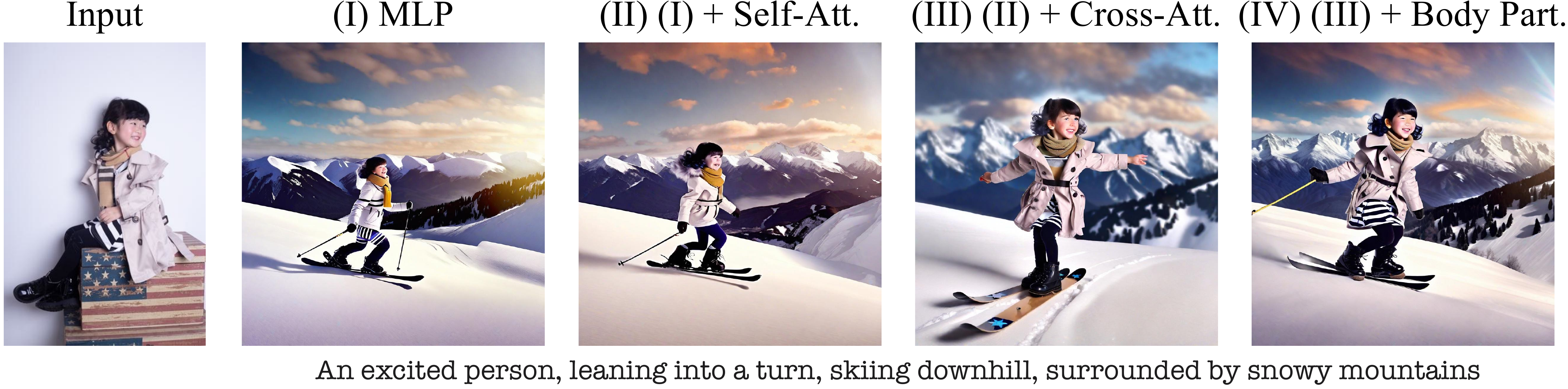} 
    \vspace{-20pt}
    \caption{\textbf{Component Analysis.}}
    \vspace{-5pt}
    
    \label{qual:appendix_abl_3}
\end{figure} 
    \begin{table}[t]
    \centering
    \resizebox{0.9
    \columnwidth}{!}{
    \begin{tabular}{c|l|ccc}
        \toprule
        & \multirow{2}{*}{Component} & \multicolumn{3}{c}{PPR10K} \\ & & D-I $\uparrow$ & D-T $\uparrow$  & D-H $\uparrow$\\
        \midrule
        (I) & MLP & 6.66 & \textbf{7.11}  & 6.74 \\
        (II) & (I) + Self-Attention 
          & 6.54 & 7.01 & 6.63 \\
        (III) & (II) + Cross-Attention & \textbf{7.47} & 6.13 & 6.40\\
        \midrule
        (IV) & (III) + Body Part Decomposition & 7.30 & 6.67 & \textbf{6.85} \\
        \bottomrule
    \end{tabular}
    }
    \vspace{-5pt}
    \caption{\textbf{Component Analysis.}}
    \label{tab:comp_analysis}
    \end{table}

\paragrapht{Detailed Comparison with StoryMaker.} In Figure~\ref{qual:comparison_storymaker} and Table~\ref{table:storymaker}, we provide detailed comparisons of Visual Persona with StoryMaker, which is a concurrent work to ours. As presented in Figure~\ref{qual:comparison_storymaker}(a) and Table~\ref{table:storymaker}(a), StoryMaker relies on reconstruction training with an unpaired dataset, which often leads to overfitting to human location, pose, and facial expressions. In contrast, our method uses cross-image training on a curated paired dataset, enabling large deformations, including pose and facial expressions, aligned with the given text. As presented in Figure~\ref{qual:comparison_storymaker}(b) and Table~\ref{table:storymaker}(b), StoryMaker encodes two-part inputs with semantic encoders and then compresses them using a resampler and a linear layer, which often lose local details in clothing and fail to disentangle different body parts. In contrast, our fine-grained decomposition and transformer encoder-decoder better preserve each part of the full-body identity. This also limits StoryMaker to top-garment Virtual Try-On (VTON), while ours supports more flexible VTON, which is further discussed in Section~\ref{sec:applications}. Additionally, as displayed in Figure~\ref{qual:comparison_storymaker}(c), StoryMaker often produces synthetic-looking outputs, possibly due to the dataset quality, while our method can generate realistic cloth textures, benefiting from our curated dataset quality.

    \begin{table}[t]
    \centering
    \resizebox{0.7
    \columnwidth}{!}{
    \begin{tabular}{l|ccc}
        \toprule
        \multirow{2}{*}{Token Length} & \multicolumn{3}{c}{PPR10K} \\ & D-I $\uparrow$ & D-T $\uparrow$ & D-H $\uparrow$ \\
        \midrule

        $l_H=4\times4$ & 5.51 & \textbf{6.90} & 5.81 \\
        $l_H=8\times8$ & 6.56 & 6.50 & 6.52 \\
        $l_H=16\times16$ (Ours) & \textbf{7.30} & 6.67 & \textbf{6.85} \\
        \bottomrule
    \end{tabular}
    }
    \vspace{-5pt}
    \caption{\textbf{Identity Embedding Token Length ($l_H$) Analysis.}}
    \vspace{-5pt}
    \label{quan:token_length}
    \end{table}

\begin{figure*}[t]
    \begin{center}
\includegraphics[width=1.0\textwidth]{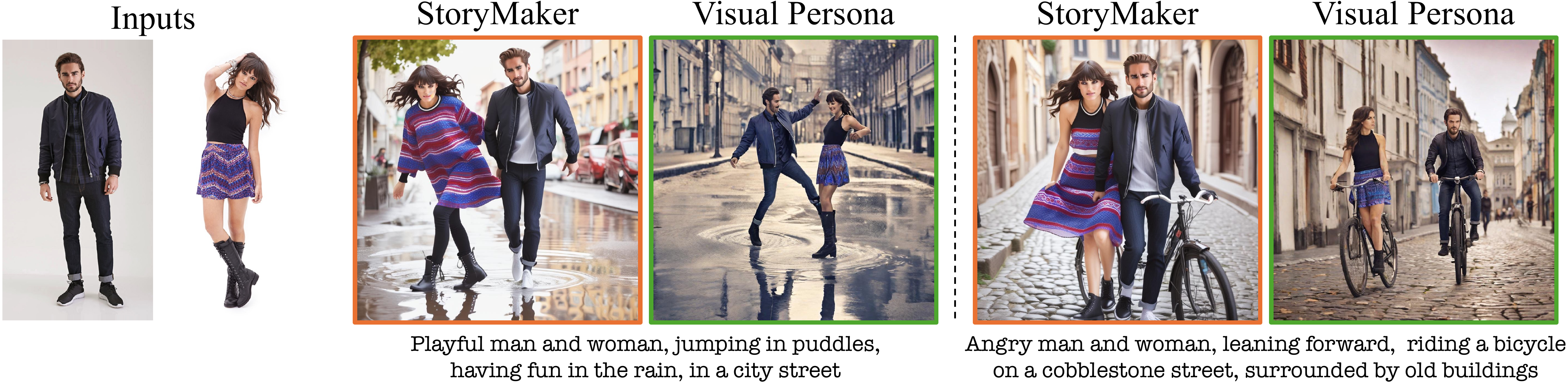} 
    \end{center}
    \vspace{-15pt}
    \caption{\textbf{Comparison for multi-person customization between StoryMaker~\cite{zhou2024storymaker} ({\color{Orange} orange}) and Visual Persona ({\color{Green} green}):} Compared to StoryMaker, Visual Persona generates more realistic interactions between multiple individuals while preserving the full-body identity of each person. Notably, Visual Persona is not trained with a multi-person dataset, as used in StoryMaker, yet our method enables multi-person customization through a simple inference modification.}
    \label{qual:comparison_MP}
\end{figure*}

\begin{figure*}[t]
    \begin{center}
\includegraphics[width=1.0\textwidth]{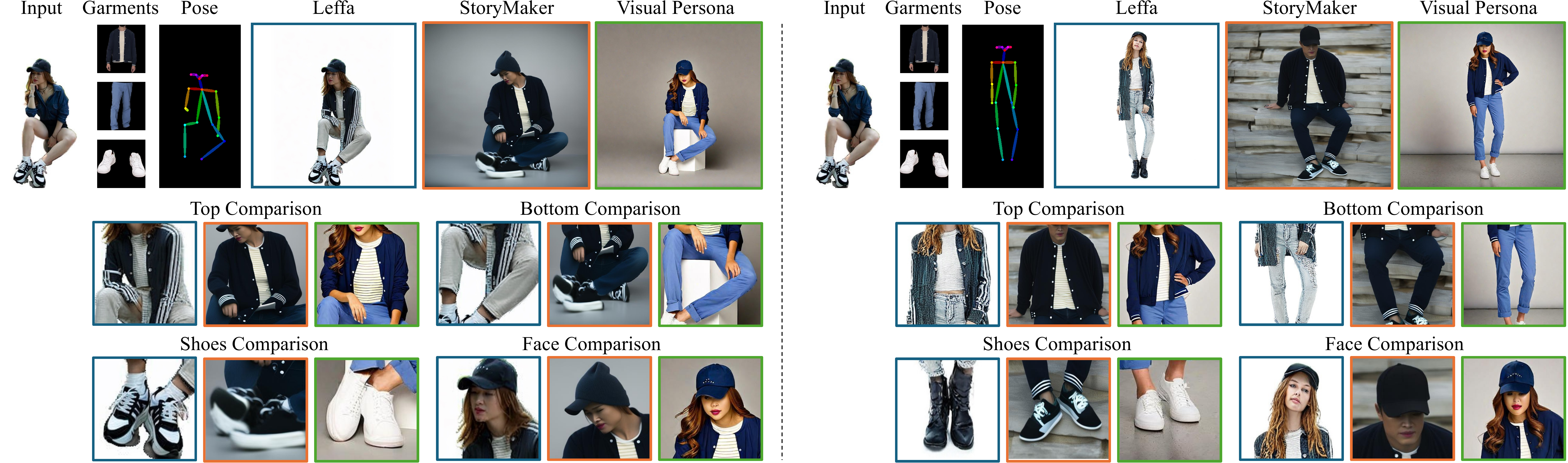} 
    \end{center}
    \vspace{-10pt}
    \caption{\textbf{Comparison for VTON between Leffa~\cite{zhou2024learning} ({\color{Blue} blue}), StoryMaker~\cite{zhou2024storymaker} ({\color{Orange} orange}), and Visual Persona ({\color{Green} green}), including full and zoomed-in images:} Compared to Leffa and StoryMaker, Visual Persona enables more flexible VTON, including top, bottom, and shoes, preserves the details of each garment, and allows accurate pose control.}
    \vspace{-10pt}
    \label{qual:comparison_VTON}
\end{figure*}

\subsection{Ablation Study}
\paragrapht{Component Analysis.} 
Figure~\ref{qual:appendix_abl_3} and Table~\ref{tab:comp_analysis} summarize the effectiveness of different configurations for the body-partitioned transformer decoder. (I) presents the results solely using an MLP layer on DINOv2~\cite{oquab2023dinov2} features, while (II) shows the results with the addition of a self-attention layer. (III) indicates the results of adding a cross-attention layer, which maps the features onto learnable identity embeddings. Compared to (I) and (II), which fail to condense detailed appearance from redundant input features, (III) shows that the transformer enhances identity preservation (D-I 6.66 vs. 7.47) but sacrifices text alignment (D-T 7.11 vs. 6.13). Compared to (III), (IV) demonstrates that body part decomposition retains the transformer's identity preservation (D-I 7.47 vs. 7.30) while significantly improving text alignment (D-T 6.13 vs. 6.67), achieving the highest D-H (6.85). This indicates that breaking down body regions is essential for the diffusion model to attend to each part separately, enabling diverse body structures in customized images.
 
\paragrapht{Identity Embedding Token Length Analysis.} Table~\ref{quan:token_length} presents an ablation study on the identity embedding token length ($l_H$), showing that increasing token length proportionally enhances identity preservation (D-I) while keeping text alignment (D-T) consistent, ultimately improving the harmonic mean (D-H). This underscores the importance of dense identity embeddings for full-body identity preservation, setting our approach apart from previous works~\cite{ye2023ip, li2024photomaker, gal2023encoder, xiao2023fastcomposer, guo2024pulid, peng2024portraitbooth, zhou2024storymaker} that compress input features into small token embeddings, typically with $l_H=16$.

\begin{figure*}[t]
    \begin{center}
\includegraphics[width=1\textwidth]{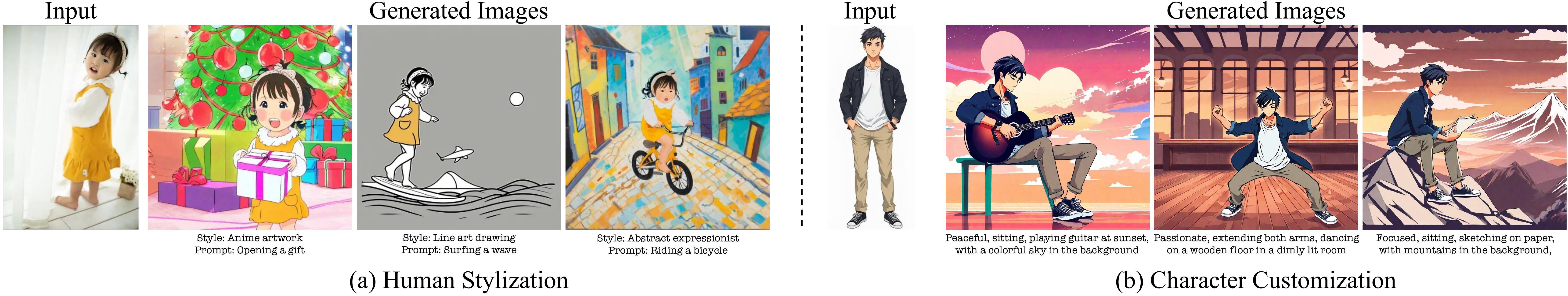} 
    \end{center}
    \vspace{-15pt}
    \caption{\textbf{Human Stylization and Character Customization.}}
    \vspace{-10pt}
    \label{qual:main_application}
\end{figure*}

\begin{figure}[t]
    \centering
    \includegraphics[width=1.0\linewidth]{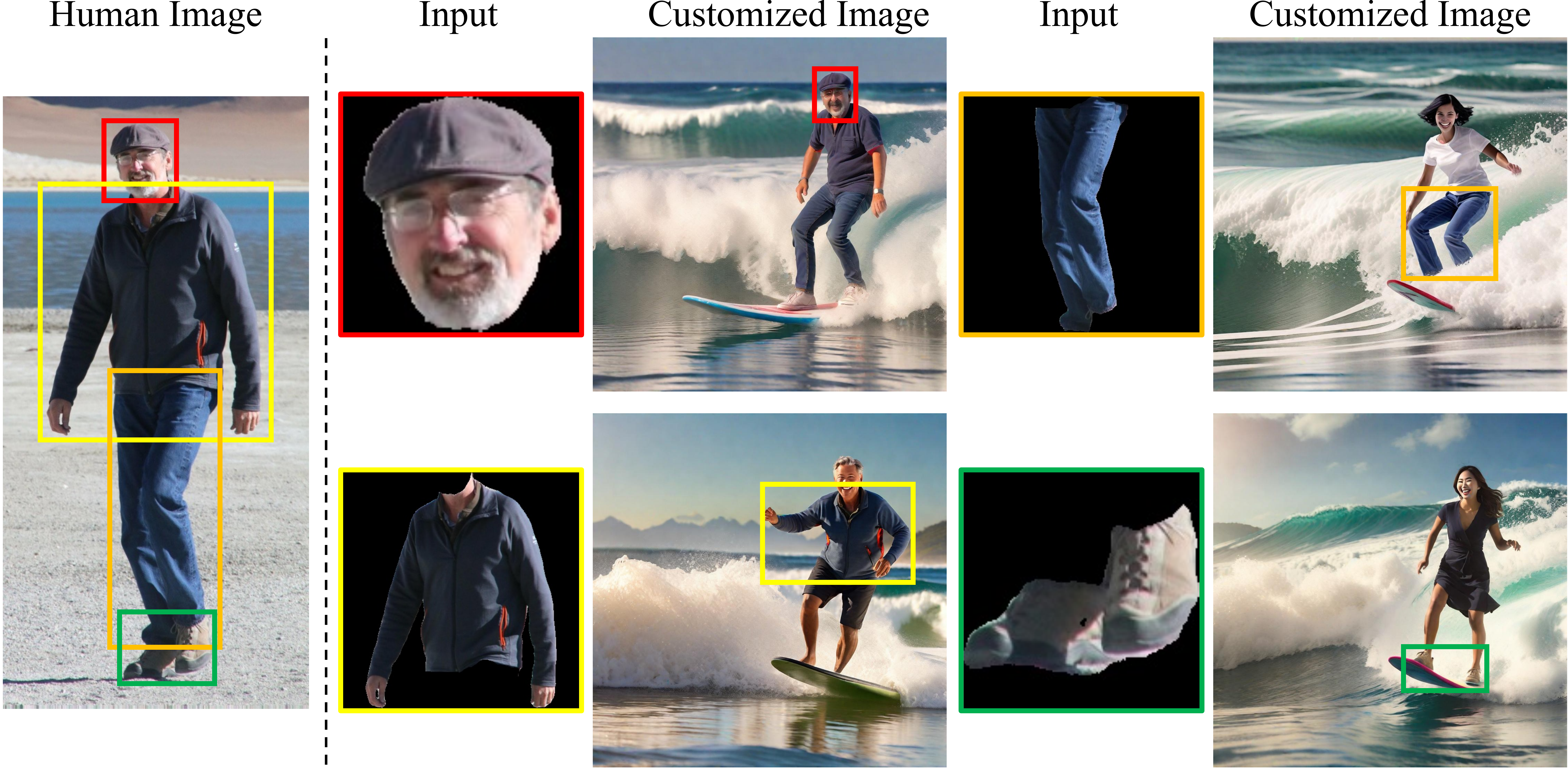} 
    \vspace{-10pt}
    \caption{\textbf{Part-Guided Full-Body Generation:} Users can select a single body part from the human image as input, allowing the pre-trained T2I diffusion model to synthesize the remaining body parts, without requiring additional training.}
    \vspace{-10pt}
    
    \label{qual:appendix_abl_2}
\end{figure}

\subsection{Application} 
\label{sec:applications}
\paragrapht{Multi-Person Customization.}
Figure~\ref{qual:comparison_MP} shows that Visual Persona supports multi-person customization without requiring the additional multi-person training used by StoryMaker~\cite{zhou2024storymaker}. This is achieved through a simple inference modification, which involves concatenating identity embeddings from multiple inputs, extracting foreground masks for each individual using text cross-attention, and augmenting identity cross-attention with these masks. StoryMaker struggles to generate interactions between multiple individuals (e.g., eye contact between two people). This is because StoryMaker is trained in a reconstruction manner, which often leads to overfitting identity-unrelated attributes from the input images (e.g., face pose, body pose, facial expression) and results in foreground-biased outputs. In contrast, Visual Persona employs cross-image training to mitigate overfitting, producing natural interactions between individuals, seamlessly integrated into the generated scenes. Additionally, StoryMaker often fails to accurately preserve the full-body appearance of each individual, while Visual Persona better retains them, benefiting from the proposed transformer architecture.

\paragrapht{Virtual Try-On (VTON).} Although not designed for VTON, Figure~\ref{qual:teaser}(a) demonstrates that Visual Persona naturally supports text-guided VTON, unlike existing VTON models~\cite{zhou2024learning, jiang2024fitdit, choi2024improving, he2022style, kim2024stableviton, zhu2024m, yang2024texture, ning2024picture}, which are limited to minor scene and pose changes due to the absence of text-based control. Specifically, given an input image for facial identity and additional images for garments, we apply a body parsing model~\cite{li2020self} to segment the face and garments and use them as inputs to our model. In this experiment, full-body images are not used.

In Figure~\ref{qual:comparison_VTON}, we also compare our method with Leffa~\cite{zhou2024learning}, the state-of-the-art VTON approach, and StoryMaker~\cite{zhou2024storymaker}. Leffa supports only top and bottom garments and requires sequential processing, which often blends garment identities. StoryMaker supports only top garments, as it decomposes the input into only two parts, the face and the whole body. Additionally, StoryMaker often struggles with pose changes and face identity preservation. In contrast, Visual Persona enables fine-grained VTON with parallel body part decomposition and better preserves full-body identity under large pose variations, benefiting from cross-image training and a transformer architecture.

\paragrapht{Human Stylization.}
Figure~\ref{qual:teaser}(b) and Figure~\ref{qual:main_application}(a) display human stylization results based on text prompts by our Visual Persona, effectively altering the image style while maintaining the full-body appearance. 

\paragrapht{Character Customization.}
Figure~\ref{qual:teaser}(c) and Figure~\ref{qual:main_application}(b) showcase the robustness of Visual Persona with out-of-domain inputs (e.g., animation domain) not included in the training set, successfully producing visually consistent outputs for anime-style input.

\paragrapht{Part-Guided Full-Body Generation.} Figure~\ref{qual:appendix_abl_2} illustrates qualitative results for part-guided full-body generation. In this experiment, we use only one body part image from the given human image as input and allow the pre-trained T2I diffusion model to synthesize the remaining body parts. The results demonstrate that Visual Persona effectively generates diverse human images while preserving the given body part, without requiring additional training, suggesting future applications such as fashion advertisements.

\begin{figure}[t]
    \centering
    \includegraphics[width=1.0\linewidth]{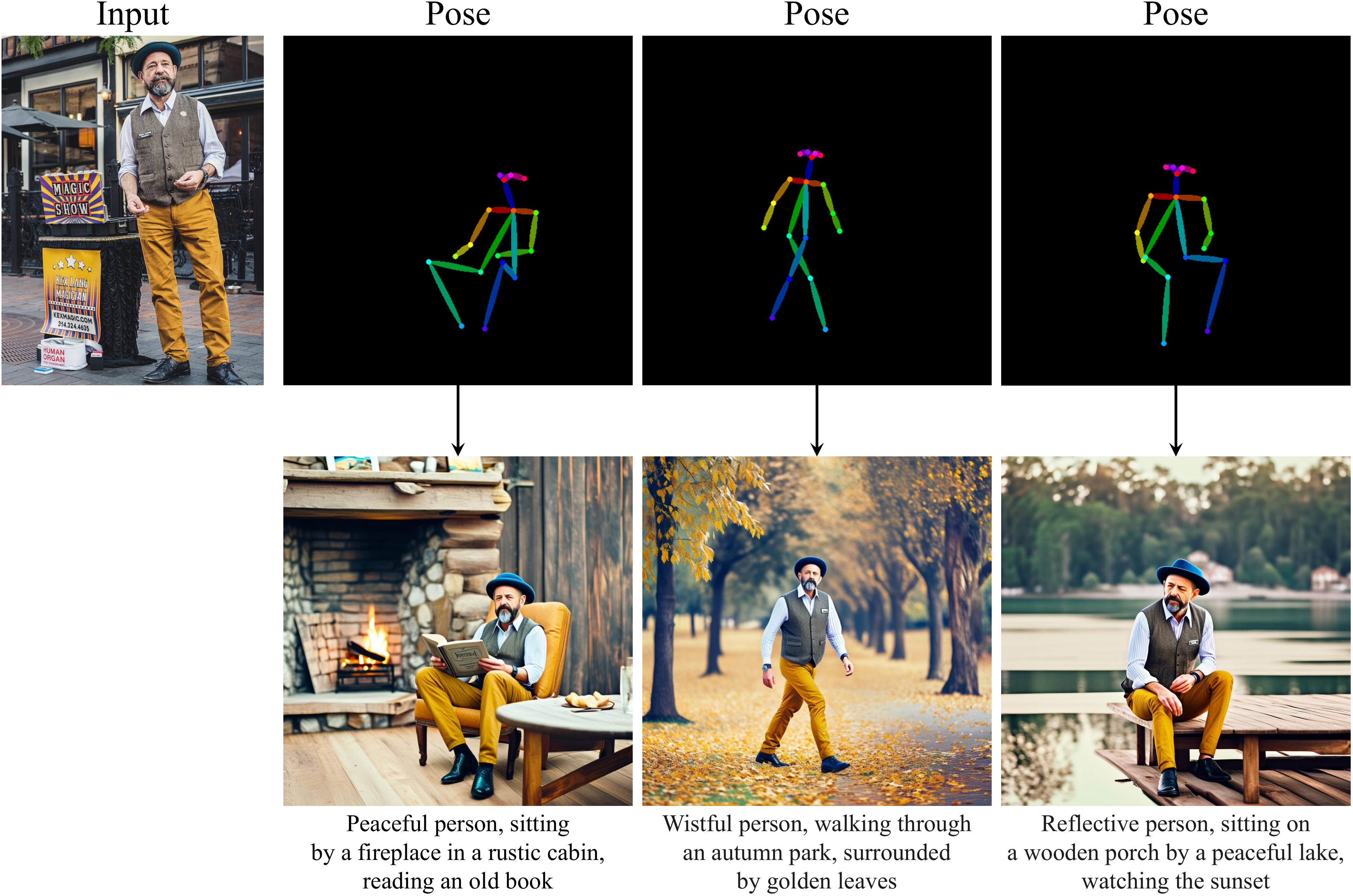} 
    \vspace{-15pt}
   \caption{\textbf{Pose-Guided Consistent Story Generation:} Users can generate a consistent story for a given human, guided by an external human pose using ControlNet~\cite{zhang2023adding}.}
    \vspace{-10pt}
    
    \label{qual:appendix_abl_4}
\end{figure} 
\paragrapht{Pose-Guided Consistent Story Generation.} Figure~\ref{qual:appendix_abl_4} illustrates the consistent story generation of a given human, following the narrative text and guided by the human pose using ControlNet~\cite{zhang2023adding}. This further highlights the practicality of our method in film production~\cite{long2024videodrafter, jin2024appearance} or book illustration~\cite{tewel2024training, akdemir2024oracle}.

\section{Conclusion}
\vspace{-5pt}
\label{sec:conclusion} 
In this paper, we introduce a foundation model for full-body human customization, dubbed \textbf{Visual Persona}. To address the difficulty in obtaining paired human datasets, we propose a data curation pipeline to collect paired human data with full-body consistency from a large pool of unpaired human images. We further introduce a transformer encoder-decoder architecture, adapted to a pre-trained T2I diffusion model, which enables precise visual transfer by projecting detailed full-body appearance into dense identity embeddings that guide the T2I model in generating customized images. Combined, Visual Persona surpasses current state-of-the-art customization methods in both GPT-based and human evaluations. We also highlight the versatility of our method across various downstream tasks.

{
    \small
    \bibliographystyle{ieeenat_fullname}
    \bibliography{main.bib}
}

 \clearpage
 \appendix
\renewcommand{\thepage}{A.\arabic{page}}
\renewcommand{\thesection}{\Alph{section}}
\renewcommand{\thesubsection}{\Alph{section}.\arabic{subsection}}
\renewcommand{\thefigure}{A.\arabic{figure}}
\renewcommand{\thetable}{A.\arabic{table}}
\setcounter{page}{1}
\setcounter{figure}{0}
\setcounter{table}{0}
\maketitlesupplementary

In the following, we additionally explain the details of implementation in Sec.~\ref{supp:imp_details}, the detailed curation process of Visual Persona-500K in Sec.~\ref{sup:VP-500K}, and the evaluation details in Sec.~\ref{supp:evaluation}, covering comparison studies, existing metric analysis, GPT-based evaluation, human evaluation on facial expressions, and human evaluation details. Further analysis of Visual Persona is presented in Sec.~\ref{supp:analysis}. Additional qualitative results are included in Sec.~\ref{supp:more_results}, and limitations are discussed in Sec.~\ref{supp:limitation}.

\section{Implementation Details}
\label{supp:imp_details}
We used a pre-trained SDXL model~\cite{podell2023sdxl} for text-to-image generation at a resolution of 1024$\times$1024. We first trained our model in a reconstruction manner on an unpaired human dataset, using the same image for $X$ and $Y$ in Equation \textcolor{NavyBlue}{6}, for 35,000 steps with a batch size of 32 and a learning rate of $1e-4$. This was followed by fine-tuning on the 580K paired human dataset, Visual Persona-500K, using paired images for $X$ and $Y$ for 35,000 additional steps with a batch size of 8 and a learning rate of $5e-6$. For GPT-based evaluation~\cite{peng2024dreambench++}, we used GPT-4o-mini~\cite{openai2024gpt4o} for all evaluations. We set $\lambda=1$ for training and $\lambda=0.7$ for all evaluations. All experiments were conducted on 8 NVIDIA A100 GPUs using the Adam optimizer~\cite{kingma2014adam}. All input images for character customization (Figure~\ref{qual:teaser}, Figure~\ref{qual:main_application}, and Figure~\ref{qual:appendix_app_qual_2}) are AI-generated images~\cite{podell2023sdxl, deepai_text2img}.

\section{Visual Persona-500K Data Curation Details}
\label{sup:VP-500K}
\begin{figure}[t]
    \centering
\includegraphics[width=1.0\linewidth]{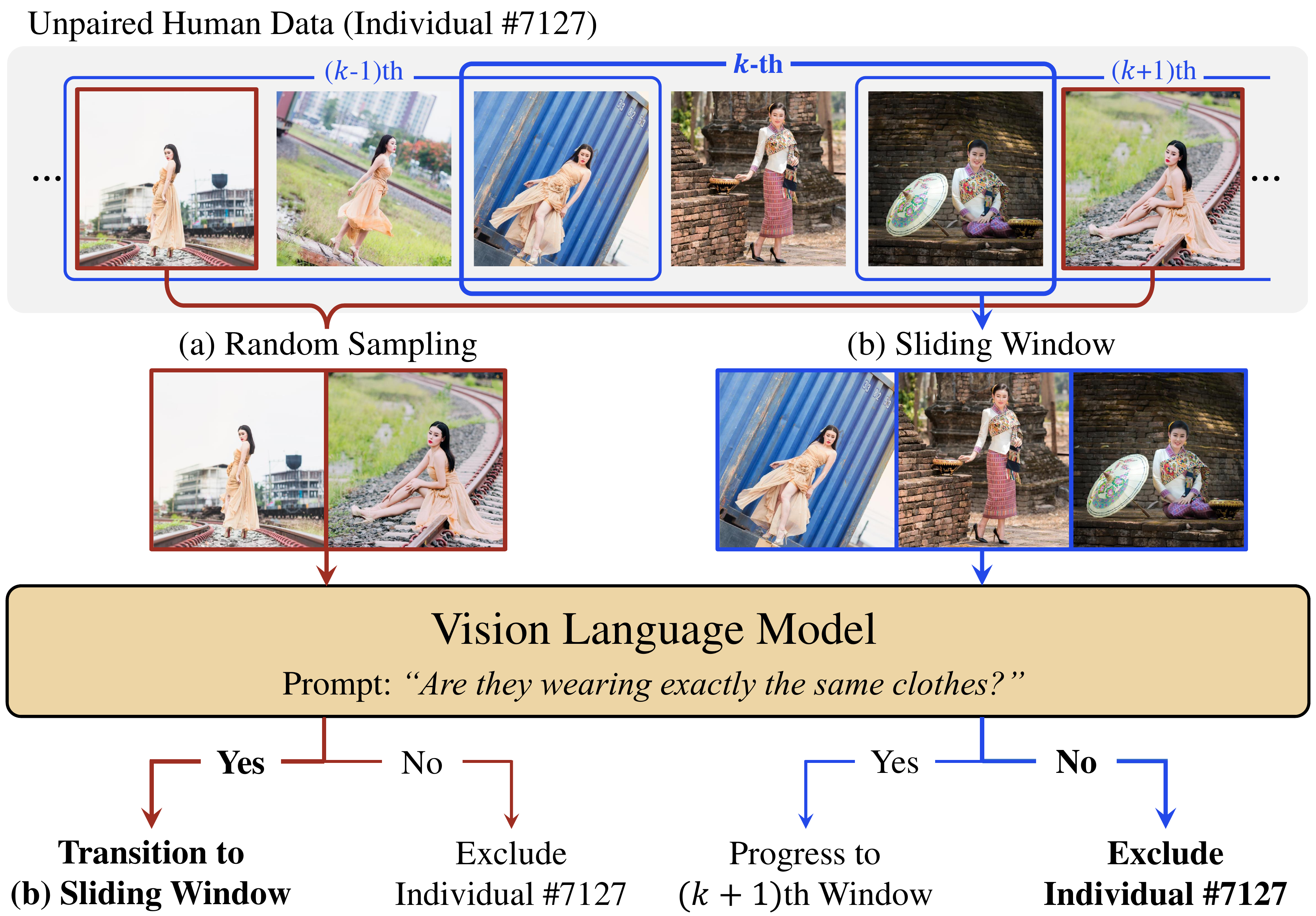} 
    \vspace{-5pt}
    \caption{\textbf{Curating Consistent Full-Body Identities.}}
    \vspace{-10pt}
    \label{qual:data_pipeline}
\end{figure}

The pipeline for curating consistent full-body identities is illustrated in Figure~\ref{qual:data_pipeline}. From the collected unpaired human pool, comprising multiple images per individual that only guarantee facial identity consistency, we aim to further evaluate body consistency using the VLM~\cite{liu2024visual}. For efficiency, we begin with two randomly selected images of the same individual and prompt LLAVA~\cite{liu2024visual} to assess whether the individual in both images is wearing the same outfit (Figure~\ref{qual:data_pipeline}(a)). A simple prompt—\textit{“Are they wearing exactly the same clothes?”}—enables the model to provide a binary decision with high precision. If the model returns a positive response, the individual is retained for further processing; otherwise, the individual is excluded from the dataset.

To ensure full-body consistency across all images for each retained individual, we further refine the dataset using a sliding window approach (Figure~\ref{qual:data_pipeline}(b)). Given our observation that LLaVA~\cite{liu2024visual} can compare up to three images, we concatenate consecutive sets of three images with a window size of 3 and a stride of 2, evaluating all sets for the same individual. If consistency is maintained across all sets, the individual is retained; otherwise, they are excluded. Ultimately, we curated a dataset of 580k paired human images across 100k unique individuals.

\section{Evaluation}
\label{supp:evaluation}
\subsection{Comparison}           
We benchmark our method against recent encoder-based zero-shot human customization models~\cite{ye2023ip, wang2024instantid, li2024photomaker, zhou2024storymaker}. These methods often focus on human face generation ({IP-Adapter-FaceID~\cite{ye2023ip}, InstantID~\cite{wang2024instantid}, PhotoMaker~\cite{li2024photomaker}}) or attempt to generate full-body images but are limited to reconstructing a single image per individual ({IP-Adapter~\cite{ye2023ip}, StoryMaker~\cite{zhou2024storymaker}}). 

Specifically, we compare the following open-source models built on SDXL~\cite{podell2023sdxl}:

\begin{itemize}
    \item \textbf{IP-Adapter-FaceID-SDXL (IP-Adapter-FaceID)}~\cite{ye2023ip}: Embeds facial features extracted from a face recognition model~\cite{deng2020retinaface, he2016deep} into small identity token embeddings, conditioning the pre-trained T2I diffusion model through a decoupled attention mechanism.
    
    \item \textbf{InstantID}~\cite{wang2024instantid}: Extends IP-Adapter-FaceID by incorporating ControlNet~\cite{zhang2023adding} to add spatial control using facial keypoints. InstantID is trained on a dataset of 50M LAION-Face~\cite{schuhmann2022laion} images and 10M face-annotated images collected internally from the web.
    
    \item \textbf{PhotoMaker}~\cite{li2024photomaker}: Stacks CLIP~\cite{radford2021learning} features from multiple face images and combines them with text embeddings to condition the T2I diffusion model. PhotoMaker is trained on a curated dataset of 112K images featuring 13K celebrities collected from the web.
    
    \item \textbf{IP-Adapter-Plus-SDXL (IP-Adapter)}~\cite{ye2023ip}: Extends the original IP-Adapter~\cite{ye2023ip} by using patch image embeddings from OpenCLIP-ViT-H-14~\cite{ilharco_gabriel_2021_5143773}.
    
    \item \textbf{StoryMaker}~\cite{zhou2024storymaker}: Combines facial features from ArcFace~\cite{deng2019arcface} and portrait features from CLIP~\cite{radford2021learning}, mapping them into small identity embeddings while fine-tuning a subset of parameters in the diffusion U-Net. StoryMaker is trained on an internally collected unpaired dataset of 500K human images, including 300K single-character and 200K two-character images. StoryMaker is concurrent work with ours.
\end{itemize}

\subsection{Dataset}
\label{supp:eval_dataset}
To evaluate our method on SSHQ~\cite{fu2022stylegan}, following its instructions, we completed the data release agreement and obtained permission for the non-commercial use of the dataset. To minimize the influence of off-the-shelf foreground mask generators~\cite{huynh2024maggie, li2020self, khirodkar2024sapiens}, we used the foreground masks provided by SSHQ~\cite{fu2022stylegan} and PPR10K~\cite{liang2021ppr10k} for evaluating all methods in this paper.

\begin{figure*}[t]
    \begin{center}
\includegraphics[width=1.0\linewidth]{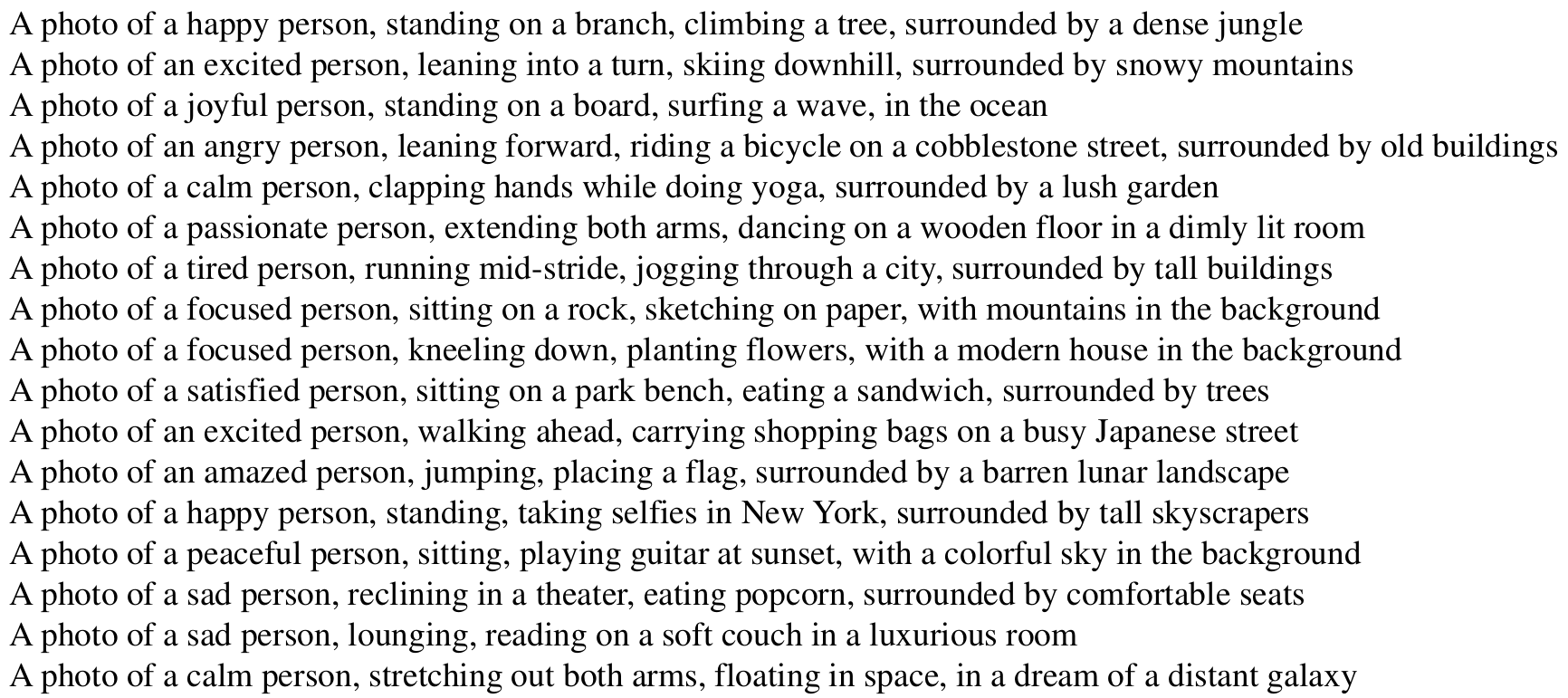} 
    \end{center}
    \vspace{-10pt}
    \caption{\textbf{Evaluation Prompts for Full-Body Human Customization:} To evaluate full-body human customization, we generated 17 text prompts by augmenting the original DreamBooth prompts~\cite{ruiz2023dreambooth} using ChatGPT~\cite{achiam2023gpt}. These prompts were utilized for all evaluations in this paper.}
    
    \label{qual:prompt}
\end{figure*}

To assess text alignment, we augmented 17 prompts for live objects in Dreambooth~\cite{ruiz2023dreambooth} using ChatGPT~\cite{achiam2023gpt}, specifically tailored for full-body human customization to follow the template \textit{“A photo of a \{facial expression\} person, \{pose\}, \{action\}, and \{surrounding\}”}. The generated prompt list is provided in Figure~\ref{qual:prompt}. This prompt list was used for all evaluations.

\begin{figure}[t]
    \centering
    \includegraphics[width=1.0\linewidth]{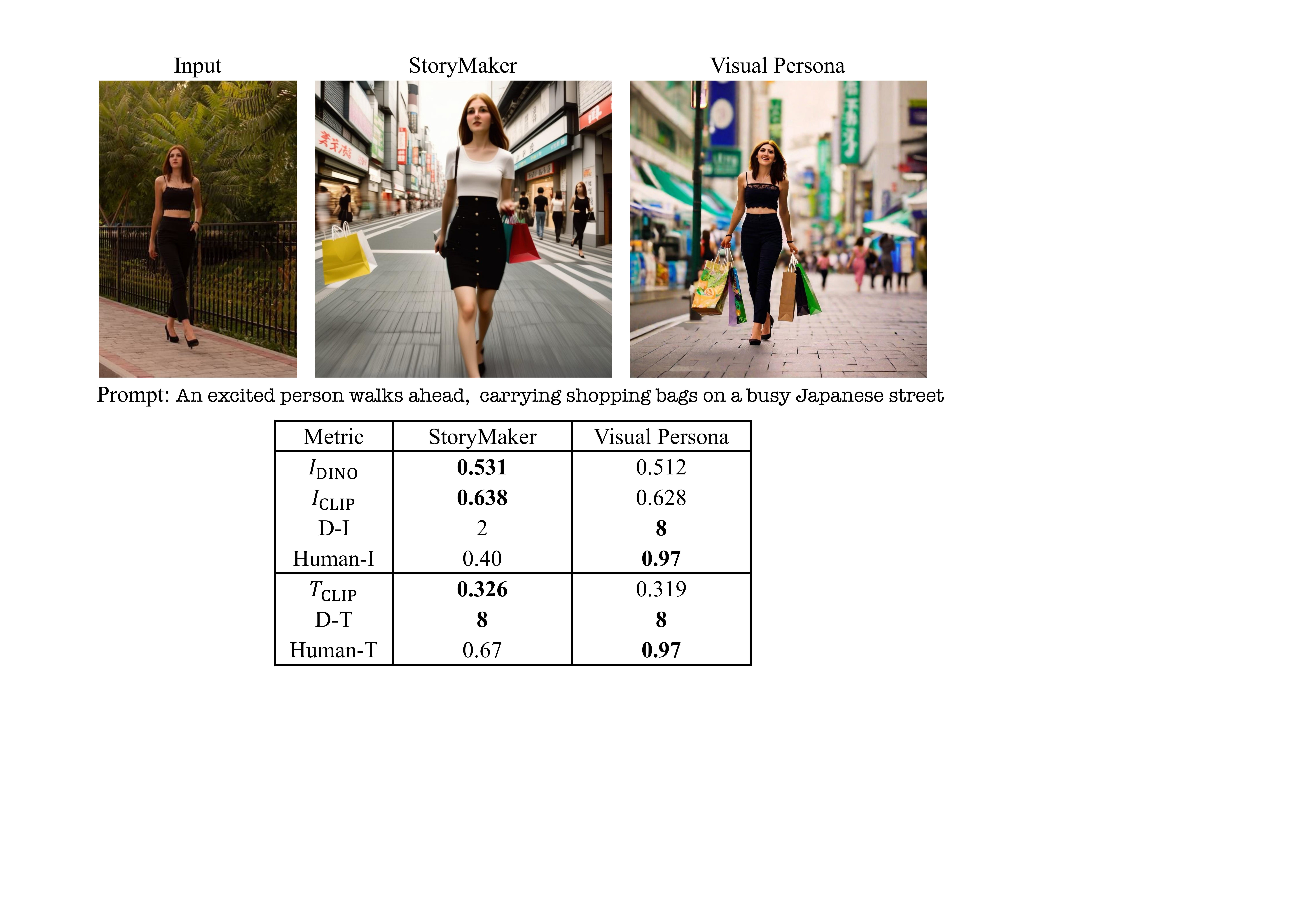} 
    \vspace{-10pt}
    \caption{\textbf{GPT-based Metrics Align Better with Human Preferences:} The upper part of the table presents evaluations for identity preservation ($I_\mathrm{DINO}$~\cite{oquab2023dinov2}, $I_\mathrm{CLIP}$~\cite{jiang2023clip}, D-I, Human-I), while the lower part presents evaluations for text alignment ($T_\mathrm{CLIP}$, D-T, Human-T). Prior metrics ($I_\mathrm{DINO}$, $I_\mathrm{CLIP}$, $T_\mathrm{CLIP}$) fail to align with human preferences (Human-I, Human-T) because they calculate cosine distances only between global feature vectors from generated images and given conditions. In contrast, GPT-based evaluations (D-I, D-T) better align with human preferences (Human-I, Human-T).
\label{qual:appendix_abl_metric}
}
    \vspace{-5pt}

\end{figure}

\subsection{Metric}
\label{supp:metrics} 
\paragrapht{GPT-based Evaluation.} As discussed in~\cite{christodoulou2024finding, tan2024evalalign, li2024genai, jiang2024genai, lin2025evaluating}, existing metrics, including identity preservation metrics, DINO image similarity ($I_\mathrm{DINO}$)~\cite{oquab2023dinov2}, CLIP image similarity ($I_\mathrm{CLIP}$)~\cite{jiang2023clip}, and the text alignment metric, CLIP image-text similarity ($T_\mathrm{CLIP}$)~\cite{jiang2023clip}, often fail to align with human preferences, struggling to accurately evaluate local appearance transfer ($I_\mathrm{DINO}$, $I_\mathrm{CLIP}$) and the alignment of complex human body structures with the given prompts ($T_\mathrm{CLIP}$). This limitation is demonstrated in Figure~\ref{qual:appendix_abl_metric}, where Visual Persona achieves higher human preference scores in identity preservation (Human-I) and text alignment (Human-T), yet existing metrics ($I_\mathrm{DINO}$, $I_\mathrm{CLIP}$, $T_\mathrm{CLIP}$) assign higher scores to StoryMaker~\cite{zhou2024storymaker} across all three metrics. This discrepancy arises because these metrics extract global vectors from the generated images and the given conditions (input image or text prompt) and calculate the distances between them, thereby ignoring local appearance details, intricate human poses, actions, and surrounding elements in the images. For human evaluation (Human-I, Human-T) in this comparison, 30 human raters were recruited to assess identity preservation and text alignment using a scale of \{0, 0.5, 1\} for not aligned, partially aligned, and fully aligned, respectively.

\begin{figure*}[t]
    \begin{center}
\includegraphics[width=0.95\linewidth]{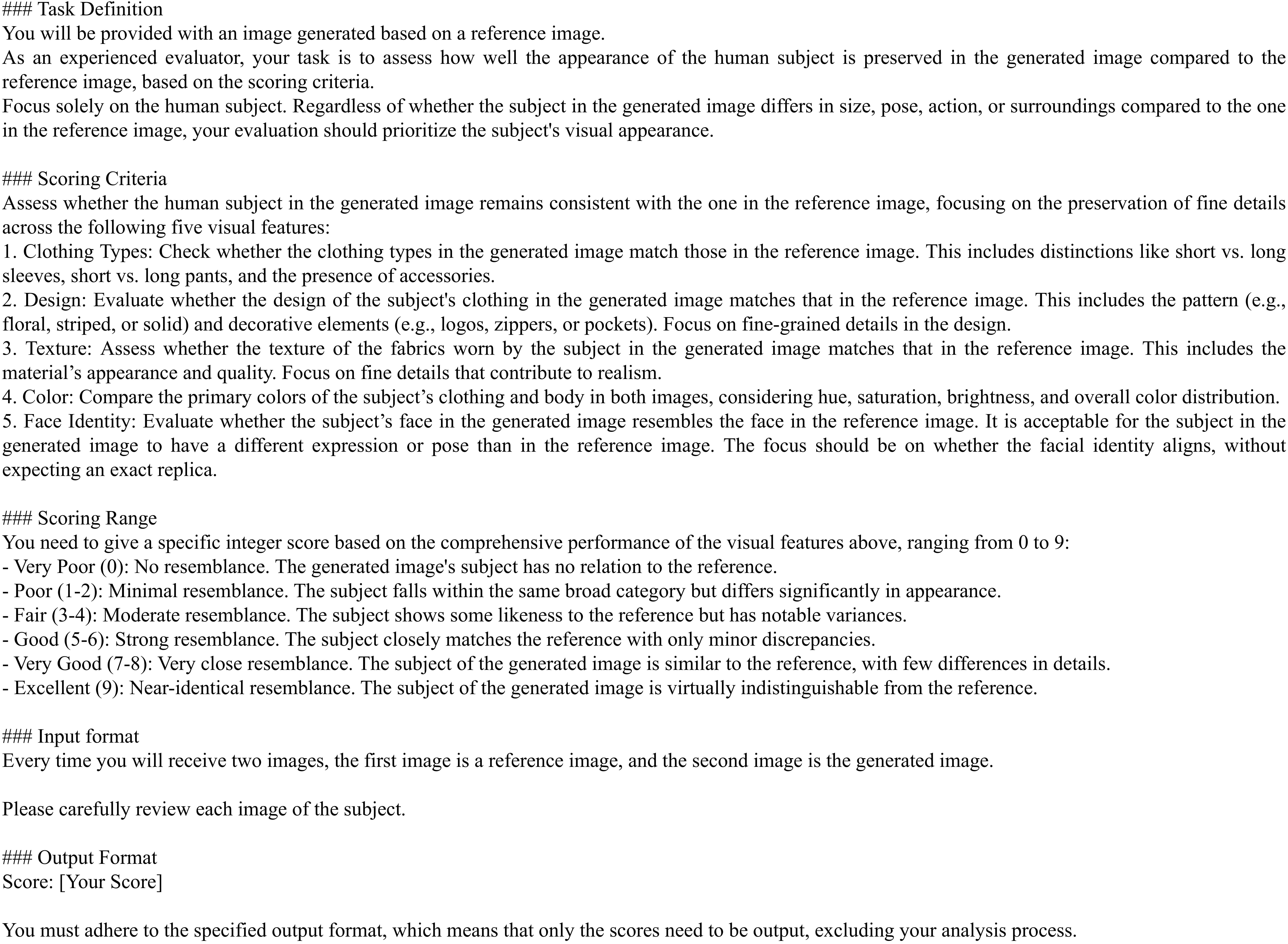} 
    \end{center}
    \vspace{-10pt}
    \caption{\textbf{User Prompts for Evaluating Identity Preservation:} To assess identity preservation, we provide evaluation instructions to GPT~\cite{openai2024gpt4o} as user prompts.}
    
    \label{qual:user_prompt_identity}
\end{figure*}

\begin{figure*}[t]
    \begin{center}
\includegraphics[width=1.0\linewidth]{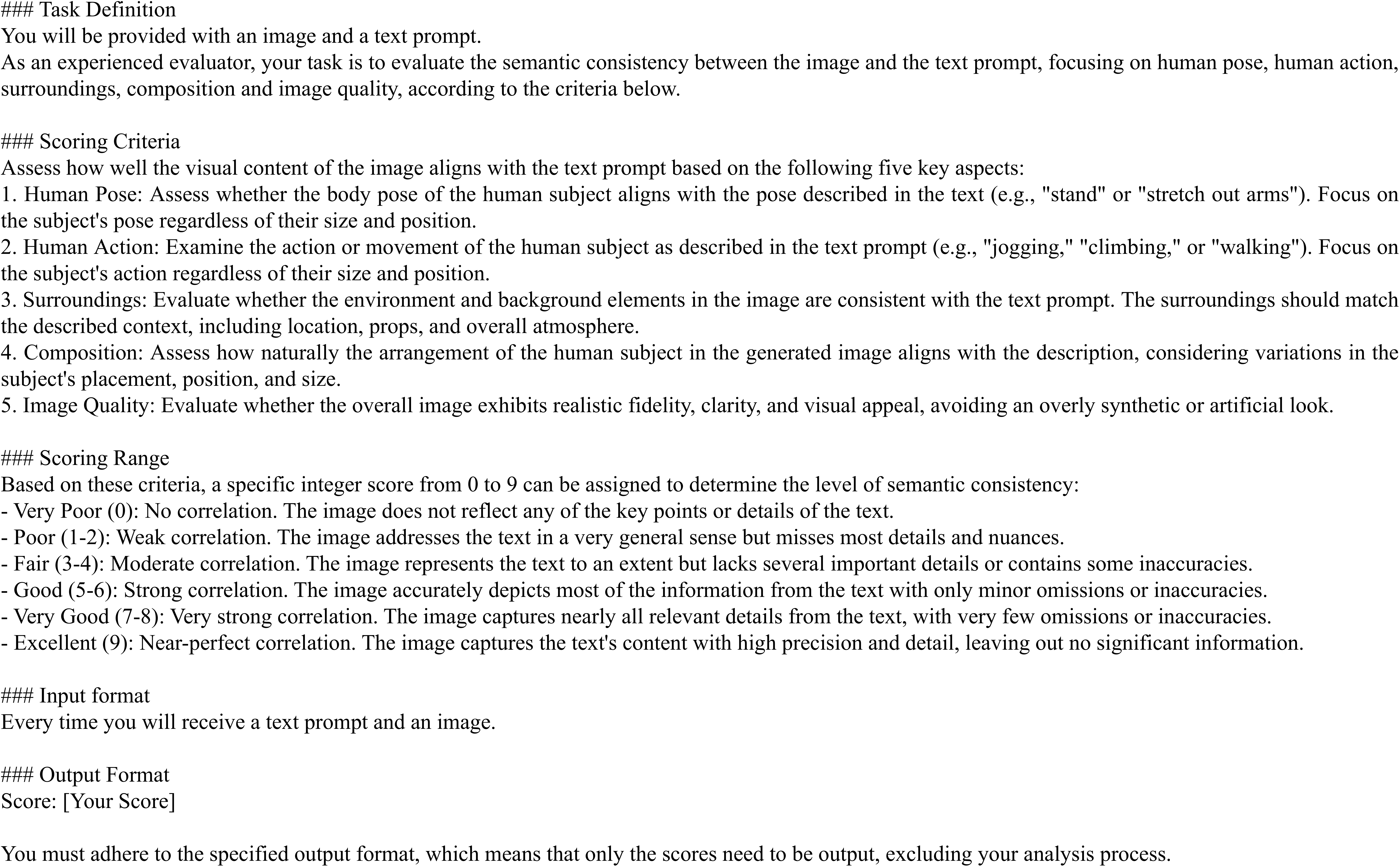} 
    \end{center}
    \vspace{-10pt}
    \caption{\textbf{User Prompts for Evaluating Text Alignment:} To assess text alignment, we provide evaluation instructions to GPT~\cite{openai2024gpt4o} as user prompts.}
    
    \label{qual:user_prompt_text}
\end{figure*}

To address this issue, we adopt Dreambench++~\cite{peng2024dreambench++}, a human-aligned, automated, GPT~\cite{achiam2023gpt}-based evaluation benchmark designed for customized image generative models. Figure~\ref{qual:appendix_abl_metric} shows that GPT-based evaluation scores for identity preservation and text alignment, denoted as D-I and D-T respectively, align more closely with human preferences compared to previous metrics. Specifically, Dreambench++~\cite{peng2024dreambench++} provides evaluation instructions as user prompts to GPT, which include the task description, scoring criteria, scoring range, and format specifications. We tailored the task description and scoring criteria for full-body human customization and adjusted the scoring range from [0, 4] to [0, 9] to enable a more comprehensive evaluation. The complete evaluation instructions for identity preservation and text alignment are provided in Figure~\ref{qual:user_prompt_identity} and Figure~\ref{qual:user_prompt_text}, respectively.

\begin{figure}[t]
    \centering
    \includegraphics[width=1.0\linewidth]{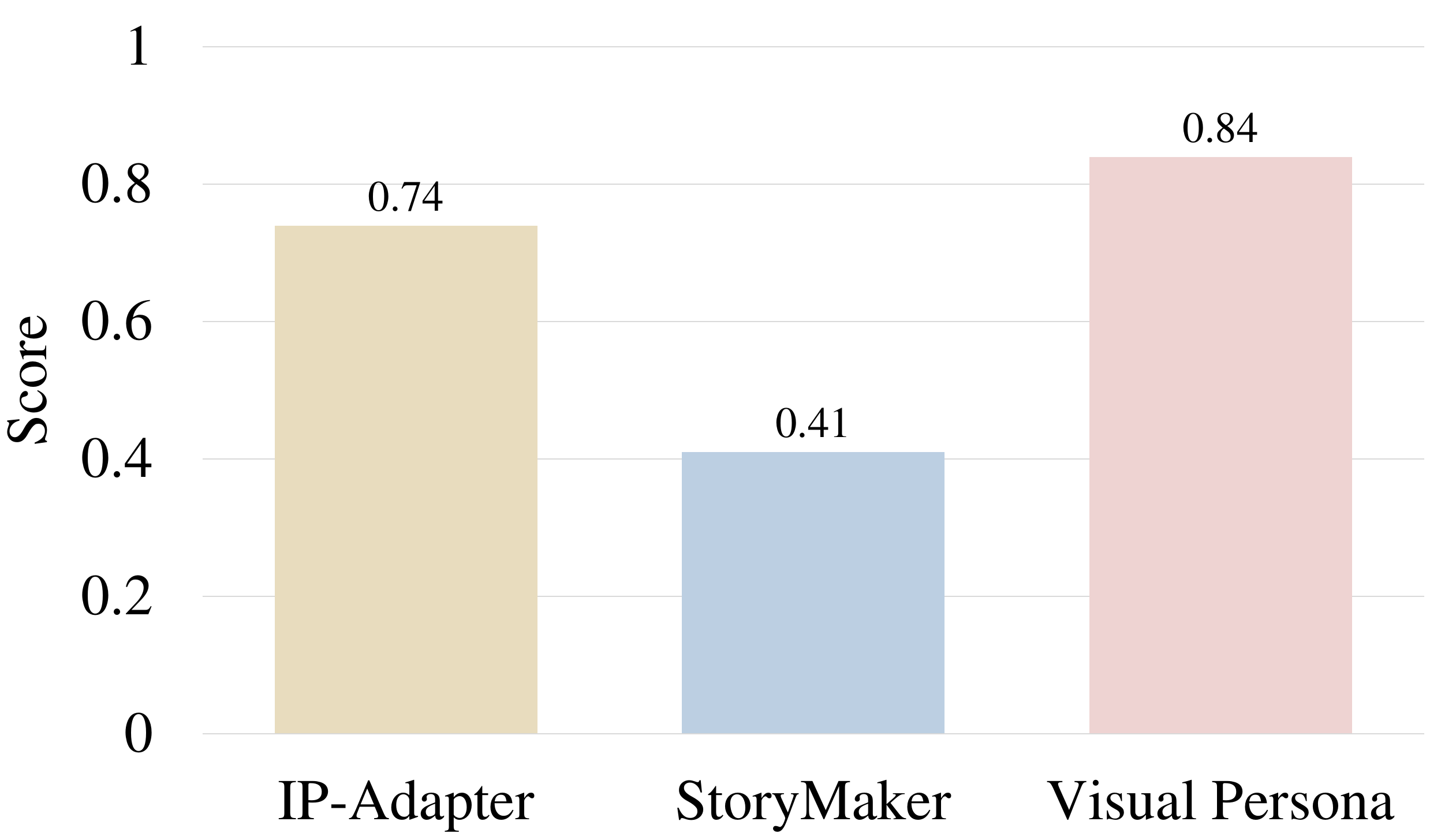} 
    \vspace{-5pt}
    \caption{\textbf{Human Evaluation on Facial Expression:} Visual Persona outperforms prior works~\cite{ye2023ip, zhou2024storymaker} in text alignment related to facial expression.}
    \vspace{-5pt}
    
    \label{qual:facial_metric}
\end{figure}

\begin{figure*}[t]
    \begin{center}
        
\includegraphics[width=0.95\linewidth]{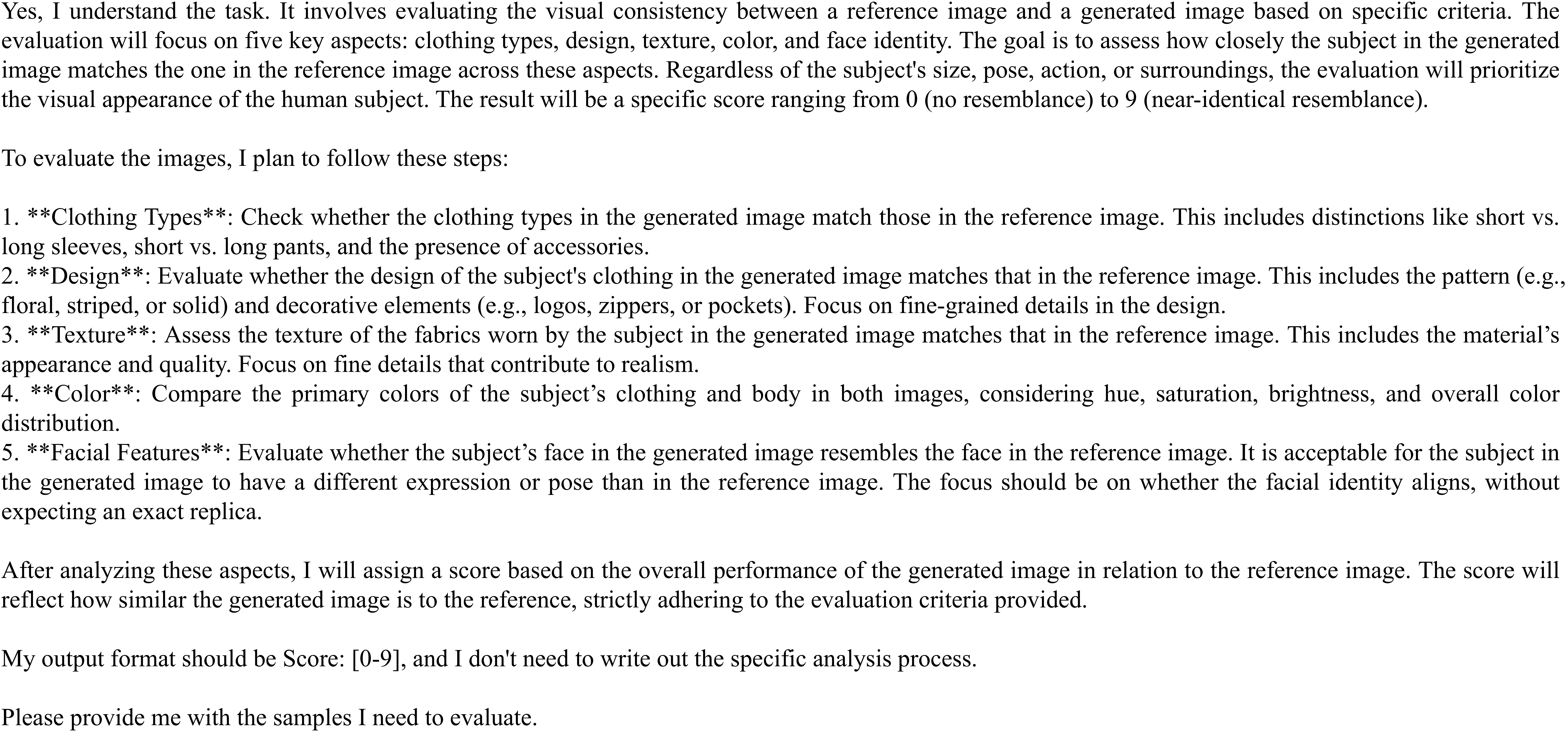} 
    \end{center}
    \vspace{-10pt}
    \caption{\textbf{Assistant Prompts for Evaluating Identity Preservation:} To assess identity preservation, we provide summary and planning responses for GPT~\cite{openai2024gpt4o} as assistant prompts.}
    
    \label{qual:gpt_prompt_identity}
\end{figure*}

\begin{figure*}[t]
    \begin{center}
\includegraphics[width=1.0\linewidth]{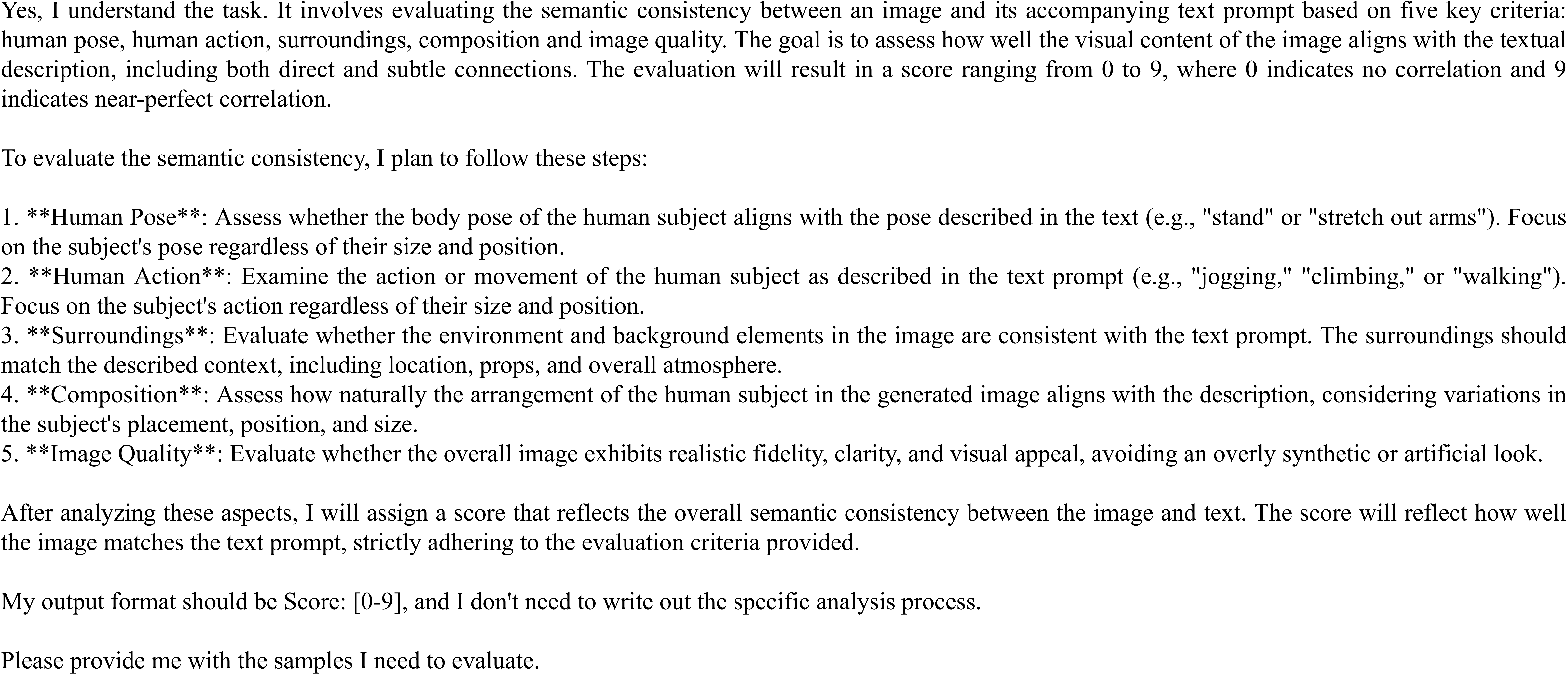} 
    \end{center}
    \vspace{-10pt}
    \caption{\textbf{Assistant Prompts for Evaluating Text Alignment:} To assess text alignment, we provide summary and planning responses to GPT~\cite{openai2024gpt4o} as assistant prompts.}
    
    \label{qual:gpt_prompt_text}
\end{figure*}

\begin{figure*}[t]
    \begin{center}
\includegraphics[width=1.0\linewidth]{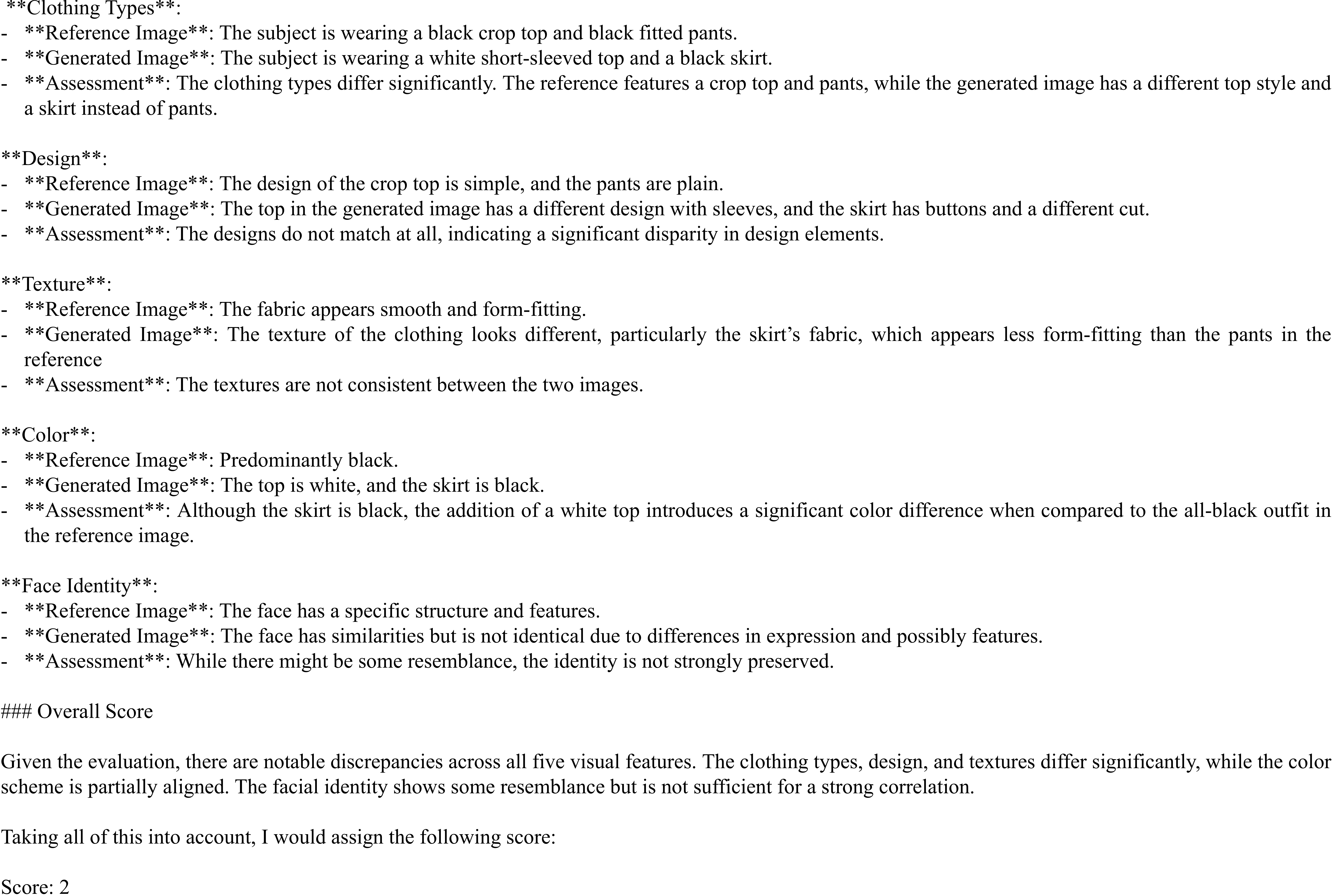} 
    \end{center}
    \vspace{-10pt}
    \caption{\textbf{Analysis Process of GPT for Evaluating Identity Preservation in StoryMaker~\cite{zhou2024storymaker} Sample (Figure~\ref{qual:appendix_abl_metric}):} GPT~\cite{openai2024gpt4o} provides a detailed analysis procedure to evaluate identity preservation based on the given scoring criteria.}
    
    \label{qual:gpt_storymaker_identity}

    \vspace{15pt}

    \begin{center}
\includegraphics[width=1\textwidth]{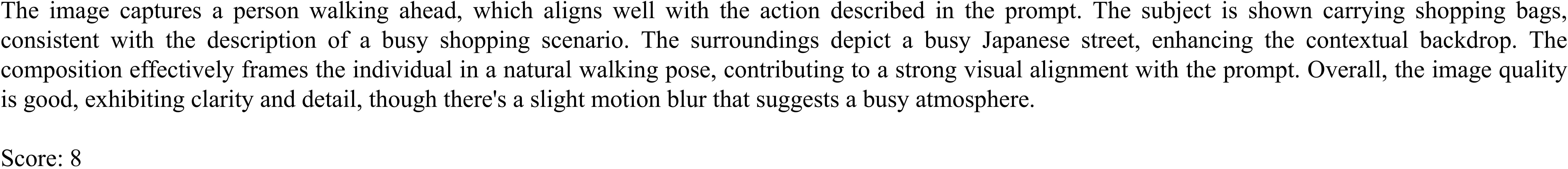} 
    \end{center}
    \vspace{-10pt}
    \caption{\textbf{Analysis Process of GPT for Evaluating Text Alignment in StoryMaker~\cite{zhou2024storymaker} Sample (Figure~\ref{qual:appendix_abl_metric}):} GPT~\cite{openai2024gpt4o} provides a detailed analysis procedure to evaluate text alignment based on the given scoring criteria.}

    \label{qual:gpt_storymaker_text}
\end{figure*}

\begin{figure*}[t]
    \begin{center}
\includegraphics[width=1.0\linewidth]{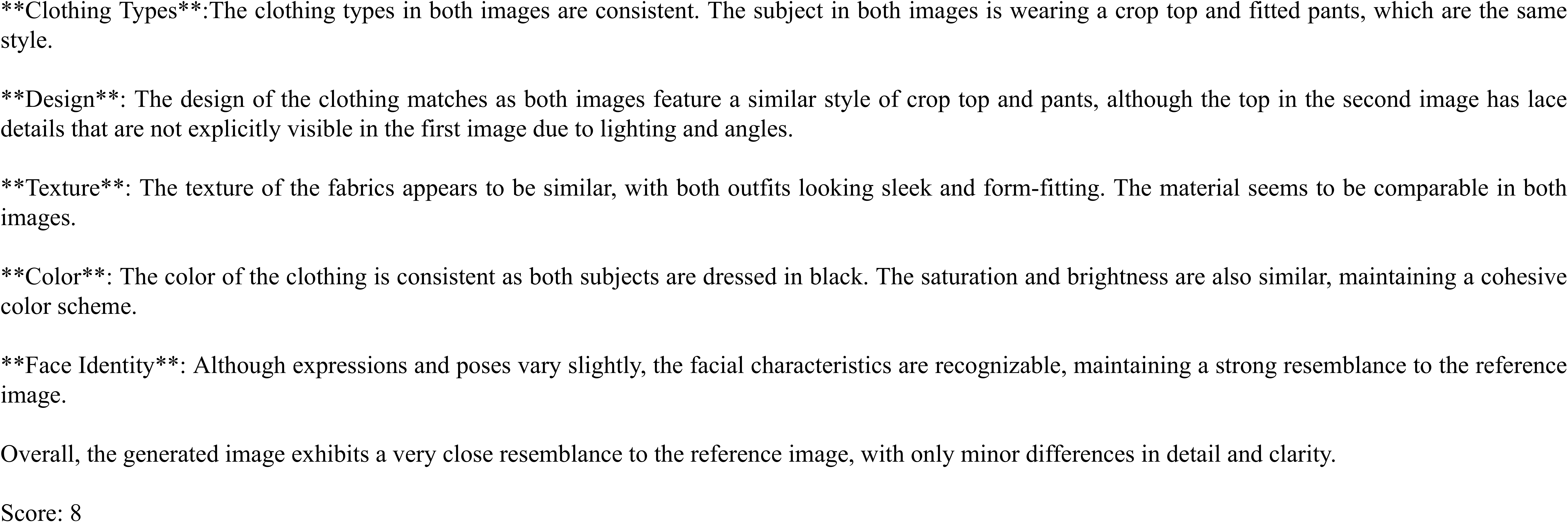} 
    \end{center}
    \vspace{-10pt}
    \caption{\textbf{Analysis Process of GPT for Evaluating Identity Preservation in Visual Persona Sample (Figure~\ref{qual:appendix_abl_metric}):} GPT~\cite{openai2024gpt4o} provides a detailed analysis procedure to evaluate identity preservation based on the given scoring criteria.}

    \label{qual:gpt_ours_identity}

    \vspace{15pt}

    \begin{center}
\includegraphics[width=1\textwidth]{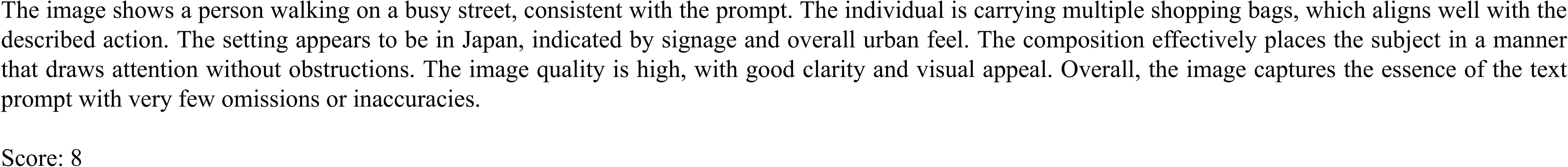} 
    \end{center}
    \vspace{-10pt}
    \caption{\textbf{Analysis Process of GPT for Evaluating Text Alignment in Visual Persona Sample (Figure~\ref{qual:appendix_abl_metric}):} GPT~\cite{openai2024gpt4o} provides a detailed analysis procedure to evaluate text alignment based on the given scoring criteria.}
    
    \label{qual:gpt_ours_text}
\end{figure*}

To align the user’s instructions with GPT’s pre-trained knowledge, Dreambench++~\cite{peng2024dreambench++} asks GPT to confirm its understanding of the task and to summarize the task itself. This process facilitates GPT's internal reasoning, enhancing task understanding and alignment with user instructions. Dreambench++ achieves this by incorporating GPT’s summary and planning responses as assistant prompts, which summarize the user instructions and outline the evaluation protocol based on the given instructions. The complete assistant prompts for identity preservation and text alignment are presented in Figure~\ref{qual:gpt_prompt_identity} and Figure~\ref{qual:gpt_prompt_text}. Note that we can further prompt GPT to output the analysis process for the scores. In Figure~\ref{qual:gpt_storymaker_identity}, ~\ref{qual:gpt_storymaker_text},~\ref{qual:gpt_ours_identity} and~\ref{qual:gpt_ours_text}, we also provide GPT's analysis procedure for evaluating the samples generated from StoryMaker~\cite{zhou2024storymaker} and Visual Persona shown in Figure~\ref{qual:appendix_abl_metric}.

\paragrapht{Human Evaluation on Facial Expression.} We observed that GPT~\cite{openai2024gpt4o} often fail to detect facial expressions when the subject is positioned far from the foreground center. To evaluate text alignment for facial expression-related prompts, we conducted a human evaluation, with the results presented in Figure~\ref{qual:facial_metric}. Eight human raters assessed whether the subject's facial expression in the generated image aligned with the given prompts. Scores were assigned as follows: 0 for no alignment, 0.5 for partial alignment, and 1 for perfect alignment.

The raters were divided into two groups, each assessing 150 images generated by three different methods: IP-Adapter~\cite{ye2023ip}, StoryMaker~\cite{zhou2024storymaker}, and Visual Persona. The same input images and prompts were used across all methods to ensure intra-rater reliability. 

Compared to StoryMaker~\cite{zhou2024storymaker}, IP-Adapter~\cite{ye2023ip} and Visual Persona demonstrate superior alignment with facial expression prompts. This difference arises because StoryMaker~\cite{zhou2024storymaker} employs ArcFace loss~\cite{deng2019arcface}, which often leads to overfitting to the pose and expression of the input image, while IP-Adapter~\cite{ye2023ip} does not account for facial expression in the text prompt during training. In contrast, Visual Persona captures facial expressions through detailed text descriptions generated by Phi-3~\cite{abdin2024phi} (Sec. \textcolor{NavyBlue}{4.1}), without relying on facial loss, enabling it to generate diverse facial expressions while maintaining facial identity consistency.

\begin{figure}[t]
    \centering
    \includegraphics[width=1.0\linewidth]{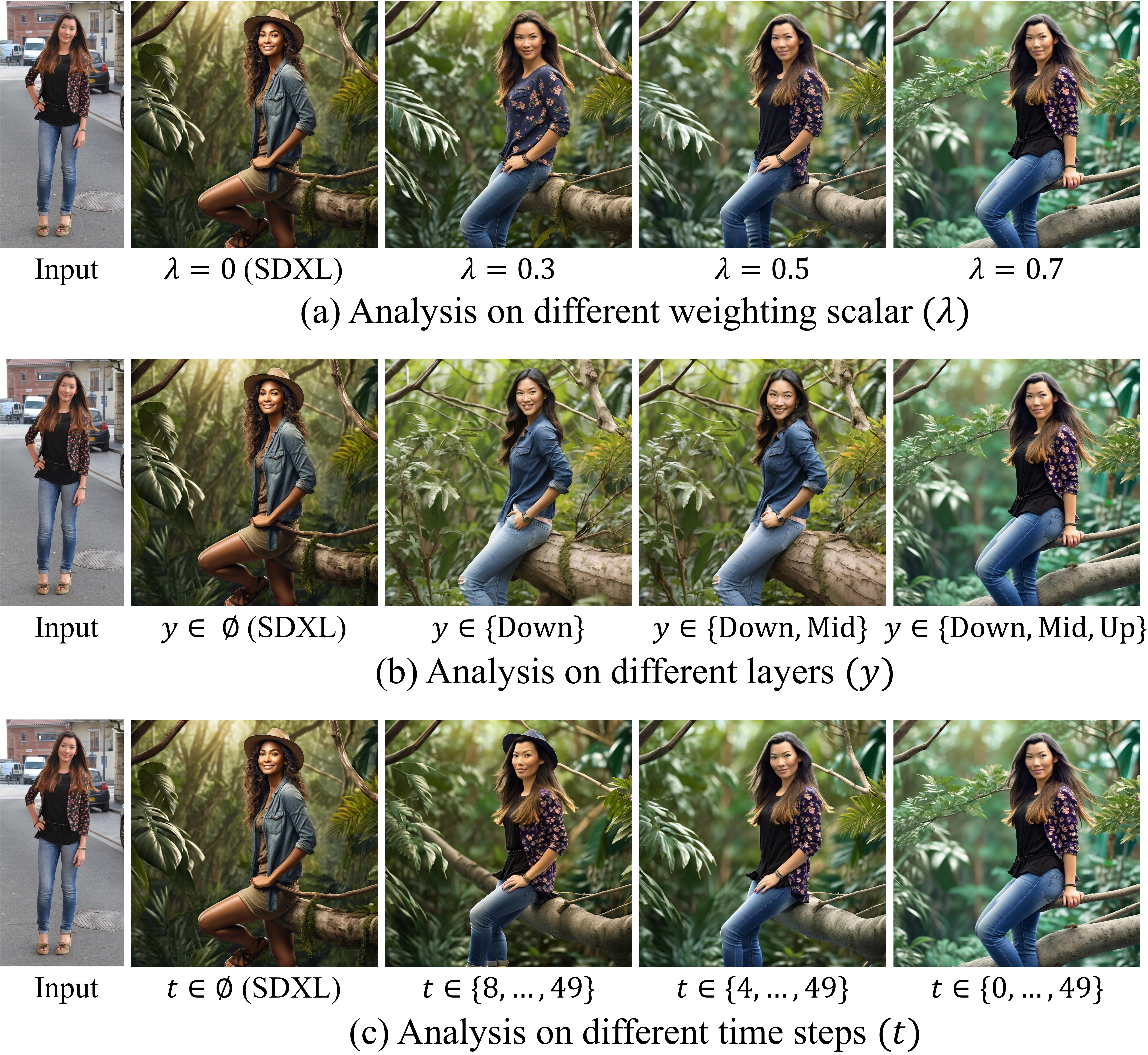} 
    \vspace{-15pt}
    \caption{\textbf{Analysis: Identity Cross-Attention Module.} Users can balance identity preservation and text alignment by adjusting the weighting scalar $\lambda$, layers $y$, and time steps $t$. 
    Increasing the weighting scalar $\lambda$ and using later layers $y$ and time steps $t$ better preserve the image structure and layout from the pre-trained SDXL~\cite{podell2023sdxl}, while slightly compromising identity preservation from the input.}
    \vspace{-5pt}
    
    \label{qual:appendix_abl}
\end{figure} 

\subsection{Human Evaluation}
\label{supp:human_evaluation}
\paragrapht{Human Evaluation Metrics.} For rigorous human evaluation, we followed the ImagenHub~\cite{ku2023imagenhub} evaluation protocol, which standardizes the assessment of conditional image generative models. ImagenHub~\cite{ku2023imagenhub} defines two human evaluation scores: Semantic Consistency ({SC}) and Perceptual Quality ({PQ}).

Semantic Consistency ({SC}) evaluates how well a generated image aligns with the provided conditions. Since our method uses an input image and a text prompt as conditions, human raters assess {SC} based on identity preservation relative to the input image and text alignment with the prompt. {SC} scores each condition independently as "inconsistent" (0 points), "partially consistent" (0.5 points), or "mostly consistent" (1 point). The final {SC} score for an image is the lowest score across the two conditions. We highlight that this metric avoids bias toward either the input image or the text prompt, as it prioritizes the lowest score, aligning with our goal of achieving both identity preservation and text alignment. Note that we applied foreground masks~\cite{fu2022stylegan, liang2021ppr10k} to the input images to assist human raters in focusing on the human parts.

Perceptual Quality ({PQ}) measures how visually convincing and natural the generated image appears, considering artifacts, distortions, and overall realism. Human raters assign 0 points if the image contains obvious artifacts or distortions, 0.5 points if the image appears unnatural with minor artifacts, and 1 point if the image looks genuine and realistic.

\paragrapht{Number of Human Raters.} Based on ImagenHub's analysis~\cite{ku2023imagenhub} showing that involving more than four human raters increases standard deviation and decreases score reliability, as measured by Krippendorff’s Alpha~\cite{hayes2007answering}, we recruited eight raters, split into two groups, each including four raters. Each group was assigned the same evaluation sheet to ensure inter-rater consistency.

\begin{figure*}[t]
    \begin{center}
\includegraphics[width=1.0\linewidth]{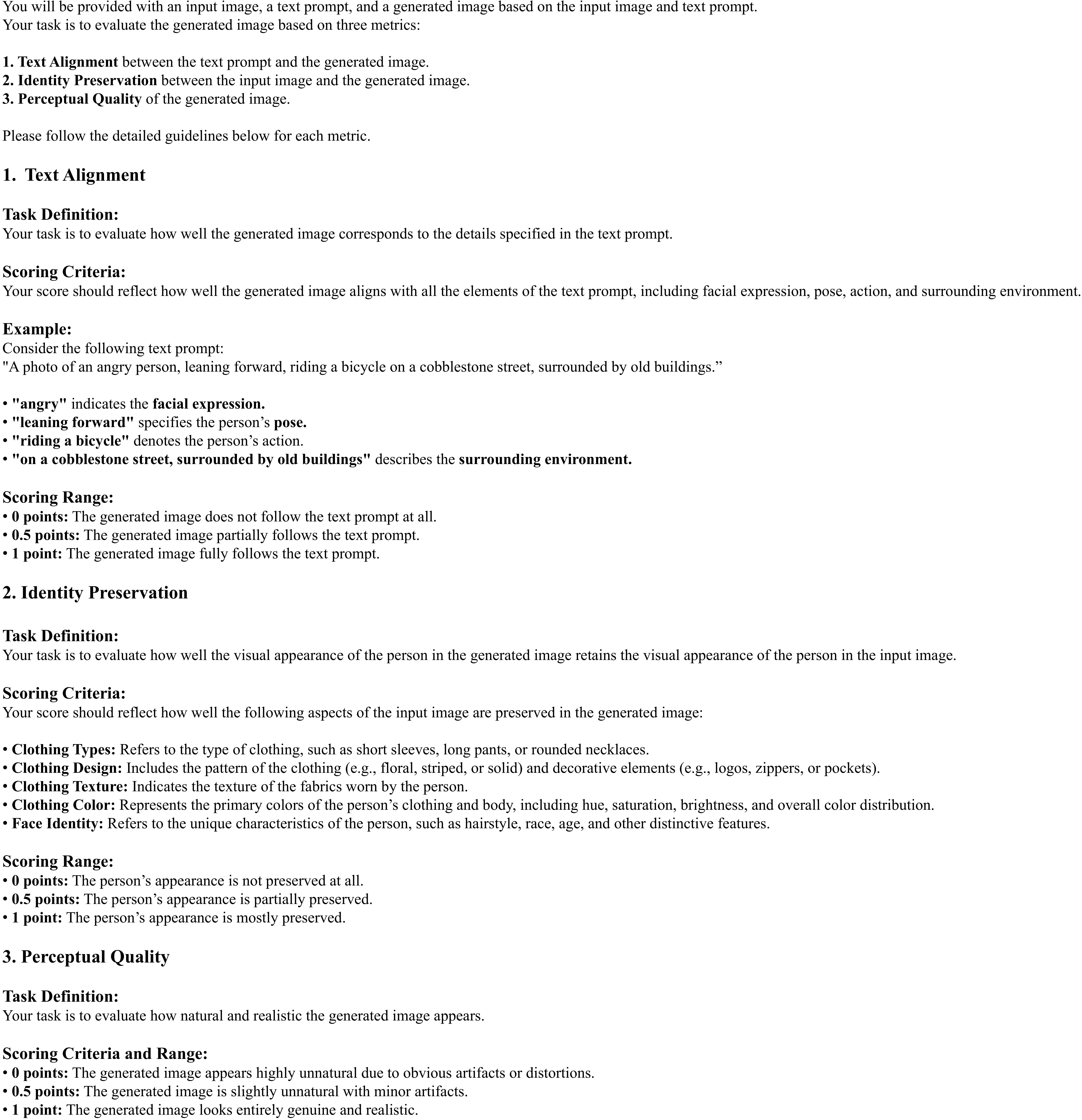} 
    \end{center}
    \vspace{-10pt}
    \caption{\textbf{Evaluation Guidelines for Human Evaluation.} We provide detailed evaluation guidelines to each human rater.}
    
    \label{qual:eval_guideline}
\end{figure*}

\begin{figure*}[t]
    \begin{center}
\includegraphics[width=0.5\linewidth]{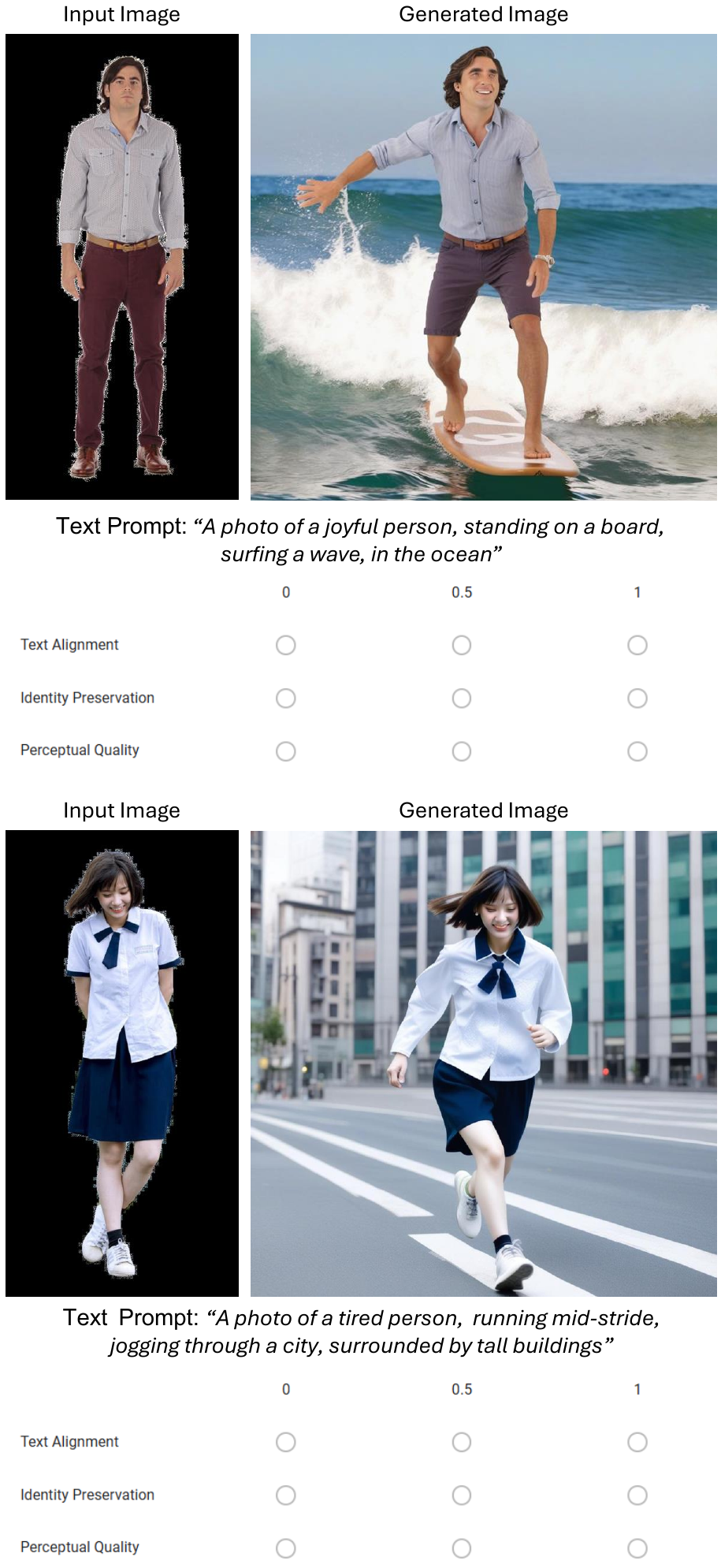} 
    \end{center}
    \vspace{-10pt}
    \caption{\textbf{Examples of Human Evaluation Questions.} }
    
    \label{qual:eval_sample}
\end{figure*}

\paragrapht{Human Evaluation Setting.} Each evaluation sheet includes 150 images generated from 50 individuals sampled from the SSHQ~\cite{fu2022stylegan} and PPR10K~\cite{liang2021ppr10k}, with 25 individuals from each dataset. We used two distinct evaluation sheets, covering a total of 300 unique images with no overlap between sheets. For each individual, one prompt was randomly sampled from the 17 pre-defined prompts, and images were generated using three different methods: IP-Adapter~\cite{ye2023ip}, StoryMaker~\cite{zhou2024storymaker}, and Visual Persona. This setup ensures intra-rater reliability, as the same rater evaluates all three methods on the same input images and text prompts. We provided detailed evaluation guidelines to the human raters. The guidelines and an example of an evaluation question are shown in Figures~\ref{qual:eval_guideline} and ~\ref{qual:eval_sample}.

\section{Analysis}
\label{supp:analysis}

\subsection{Identity Cross-Attention Module}
\paragrapht{Weighting Scalar}. Figure~\ref{qual:appendix_abl}(a) presents an ablation study on the weighting scalar $\lambda$ in Equation \textcolor{NavyBlue}{5}. $\lambda = 0$ indicates that the identity cross-attention is disabled, which is identical to the original SDXL~\cite{podell2023sdxl}. For a fair comparison, we fix the layers and time steps for identity cross-attention to include all cross-attention layers in SDXL and all 50 sampling time steps.

The results show that increasing $\lambda$ enhances identity preservation from the input but slightly degrades the original image structure in pre-trained SDXL, including background details and human pose. This suggests that users can control $\lambda$ to balance the degree of identity preservation from the input and the text alignment derived from SDXL.

\paragrapht{Layers.} 
Figure~\ref{qual:appendix_abl}(b) presents an ablation study on different layers in SDXL~\cite{podell2023sdxl}, denoted as $y$, where the identity cross-attention module is applied. Here, {Down}, {Mid}, and {Up} refer to the down blocks, mid block, and up blocks in the diffusion U-Net, respectively. For a fair comparison, we set $\lambda = 0.7$ across all 50 sampling time steps.

As discussed in~\cite{cao2023masactrl, nam2024dreammatcher}, the down blocks primarily capture image structure and layout, including background details and human pose, while the up blocks focus on image appearance. In line with this, applying identity cross-attention in the down and mid blocks significantly limits identity injection, while preserving the original image structure generated by the pre-trained SDXL. On the other hand, adding identity cross-attention to the up blocks effectively injects identity while maintaining the pre-trained SDXL image structure. Based on this observation, we utilize all cross-attention layers in the identity cross-attention module for all evaluations presented in this paper.

\paragrapht{Time Steps.} 
Figure~\ref{qual:appendix_abl}(c) shows the results of using the identity cross-attention module at different sampling time steps $t$. Since earlier time steps focus on producing image structure, while later time steps refine image appearance~\cite{cao2023masactrl, nam2024dreammatcher}, applying identity cross-attention at later time steps better preserves the original SDXL~\cite{podell2023sdxl} image structure but compromises identity preservation. In our experiment, $t \in \{4,...,49\}$ achieves strong identity preservation while maintaining the original image structure of the pre-trained SDXL. This demonstrates that users can adjust the sampling time steps to balance identity injection and generative fidelity. In our main paper, for a fair comparison with other works, we used all sampling time steps $t \in \{0,...,49\}$ for all evaluations.

\begin{table}[t]
\centering
\resizebox{1.0\columnwidth}{!}{%
\begin{tabular}{l|l|ccc|ccc}
    \toprule
     \multirow{2}{*}{Method} & \multirow{2}{*}{Body Parsing Method} & \multicolumn{3}{c|}{SSHQ} & \multicolumn{3}{c}{PPR10K} \\ 
     &  & \multicolumn{1}{c}{D-I $\uparrow$} & \multicolumn{1}{c}{D-T $\uparrow$} & \multicolumn{1}{c|}{D-H $\uparrow$} & \multicolumn{1}{c}{D-I $\uparrow$} & \multicolumn{1}{c}{D-T $\uparrow$} & \multicolumn{1}{c}{D-H $\uparrow$} \\
    \midrule
    
    StoryMaker & SSHQ~\cite{fu2022stylegan} / PPR10K~\cite{liang2021ppr10k}& 6.74 & 7.08 & 6.71 &  6.80 & 6.77 & 6.63 \\
    \midrule
    {Visual Persona} & SSHQ~\cite{fu2022stylegan} / PPR10K~\cite{liang2021ppr10k} & 7.10 & \textbf{7.15} & {6.99} & \textbf{7.30} & 6.67 & {6.85} \\
    {Visual Persona} & SCHP~\cite{li2020self} & \textbf{7.20} & \textbf{7.15}& \textbf{7.10} & {7.10} & \textbf{7.12} & \textbf{7.00} \\

    \bottomrule
\end{tabular}%
}
\vspace{-5pt}
\caption{\textbf{Analysis: Impact of Different Body Parsing Masks on Generalization.} Visual Persona remains robust regardless of the body parsing masks used.}
\vspace{-10pt}
\label{tab:body_parsing}
\end{table}

\subsection{Robustness to Model Components} Visual Persona incorporates off-the-shelf model components, including CLIP~\cite{radford2021learning}, DINO~\cite{oquab2023dinov2}, and body parsing. Notably, these components are standard in text-to-image and human-centric models. In Table~\ref{tab:body_parsing}, we further demonstrate Visual Persona's robustness to body parsing by comparing results obtained using body parsing masks from the evaluation dataset~\cite{fu2022stylegan, liang2021ppr10k} with those generated by the off-the-shelf method SCHP~\cite{li2020self}, demonstrating that our method generalizes well across different body parsing masks.

\section{More Results}
\label{supp:more_results}
\begin{figure*}[t]
    \begin{center}
\includegraphics[width=1.0\linewidth]{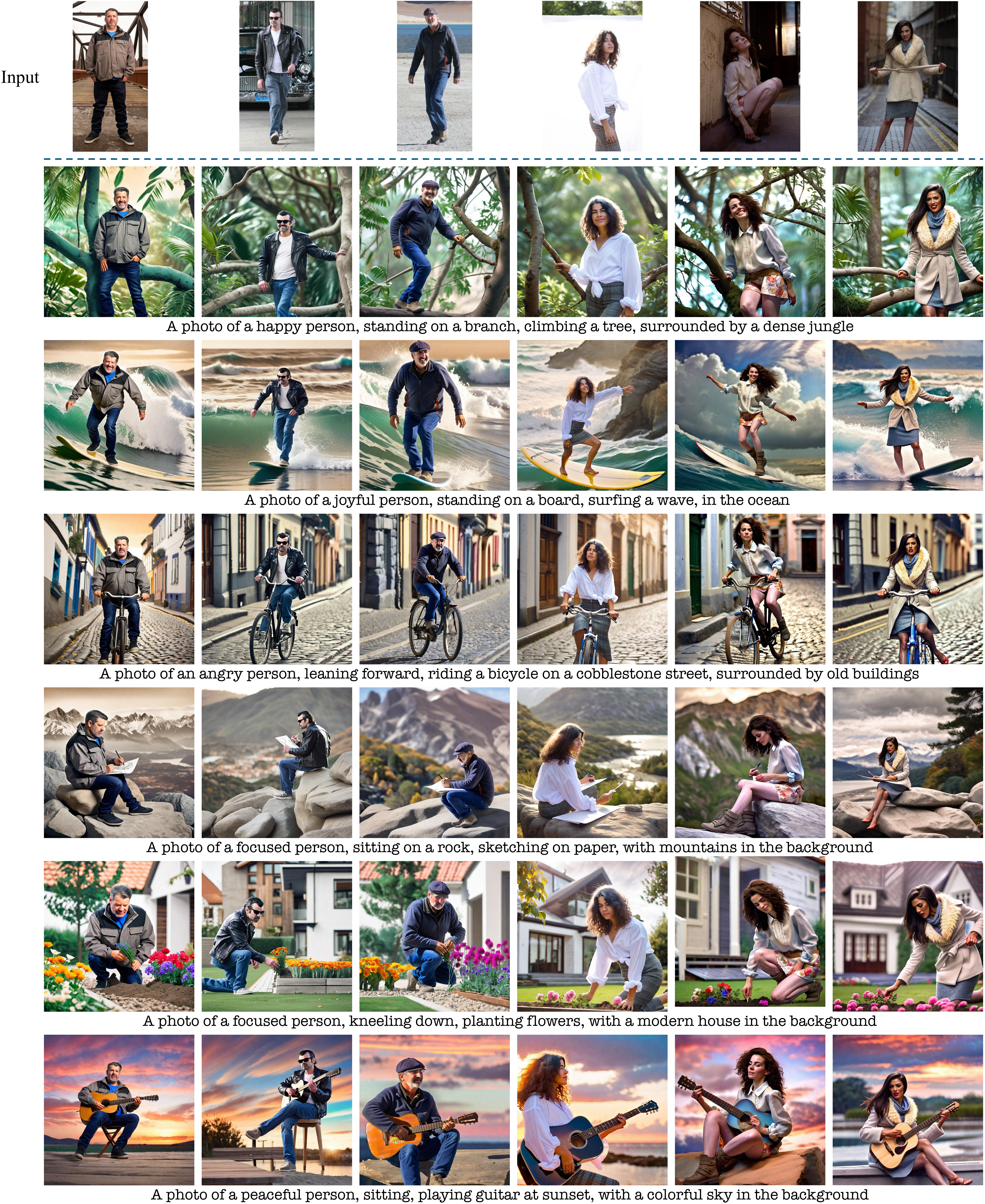} 
    \end{center}
    \vspace{-15pt}
    \caption{\textbf{Qualitative Results of Visual Persona on SSHQ~\cite{fu2022stylegan} and PPR10K~\cite{liang2021ppr10k}:} The first row includes input human images from SSHQ and PPR10K. The second to last rows include the generated images by Visual Persona based on the input images and the given prompts. Visual Persona generates full-body consistent, customized images of the input human, while closely aligning with the diverse text prompts. }

    \label{qual:appendix_qual_1}
\end{figure*}

\begin{figure*}[t]
    \begin{center}
\includegraphics[width=1.0\linewidth]{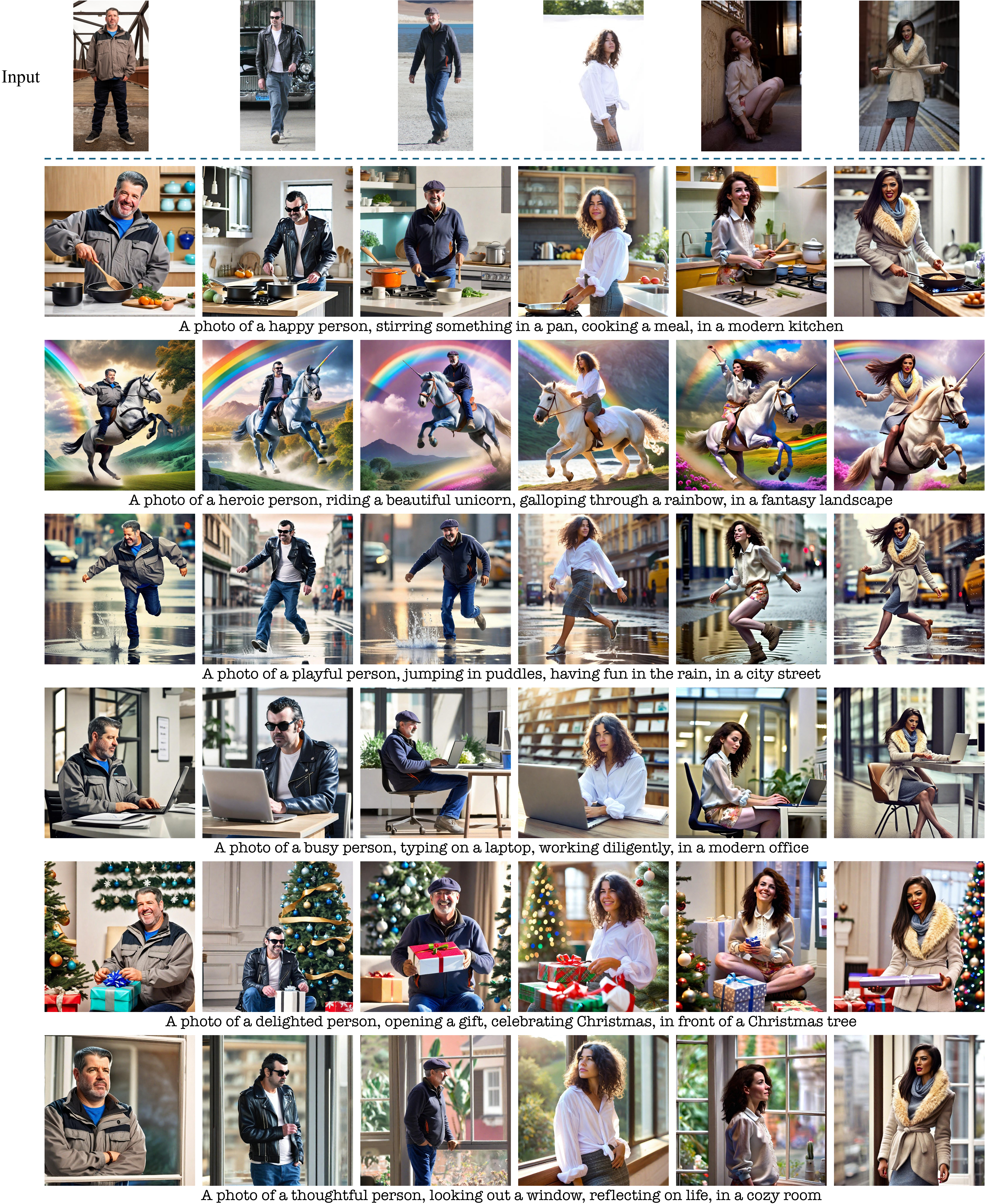} 
    \end{center}
    \vspace{-15pt}
    \caption{\textbf{Qualitative Results of Visual Persona on SSHQ~\cite{fu2022stylegan} and PPR10K~\cite{liang2021ppr10k}:} The first row includes input human images from SSHQ and PPR10K. The second to last rows include the generated images by Visual Persona based on the input images and the given prompts. Visual Persona generates full-body consistent, customized images of the input human, while closely aligning with the diverse text prompts.}

    \label{qual:appendix_qual_2}
\end{figure*}

\begin{figure*}[t]
    \begin{center}
\includegraphics[width=1.0\linewidth]{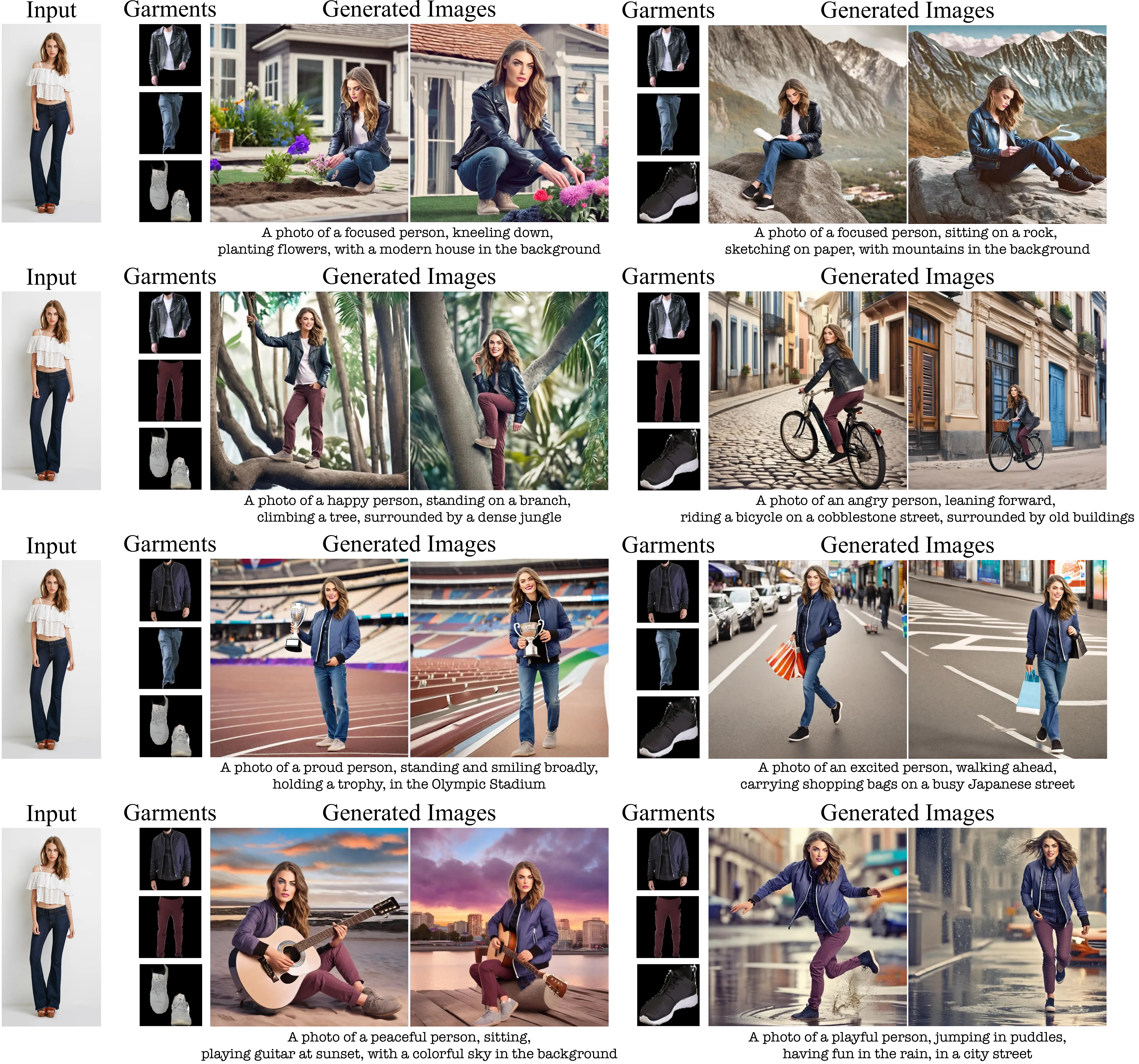} 
    \end{center}
    \vspace{-15pt}
    \caption{\textbf{Qualitative Results of Visual Persona for Text-Guided Virtual Try-On (VTON):} Although Visual Persona is not specifically designed for VTON, our method naturally supports text-guided VTON, whereas existing VTON models~\cite{zhou2024learning, jiang2024fitdit, choi2024improving, he2022style} are limited to minor scene and pose changes due to the absence of text control.}

    \label{qual:appendix_app_qual_1}
\end{figure*}

\begin{figure*}[t]
    \begin{center}
\includegraphics[width=0.7\linewidth]{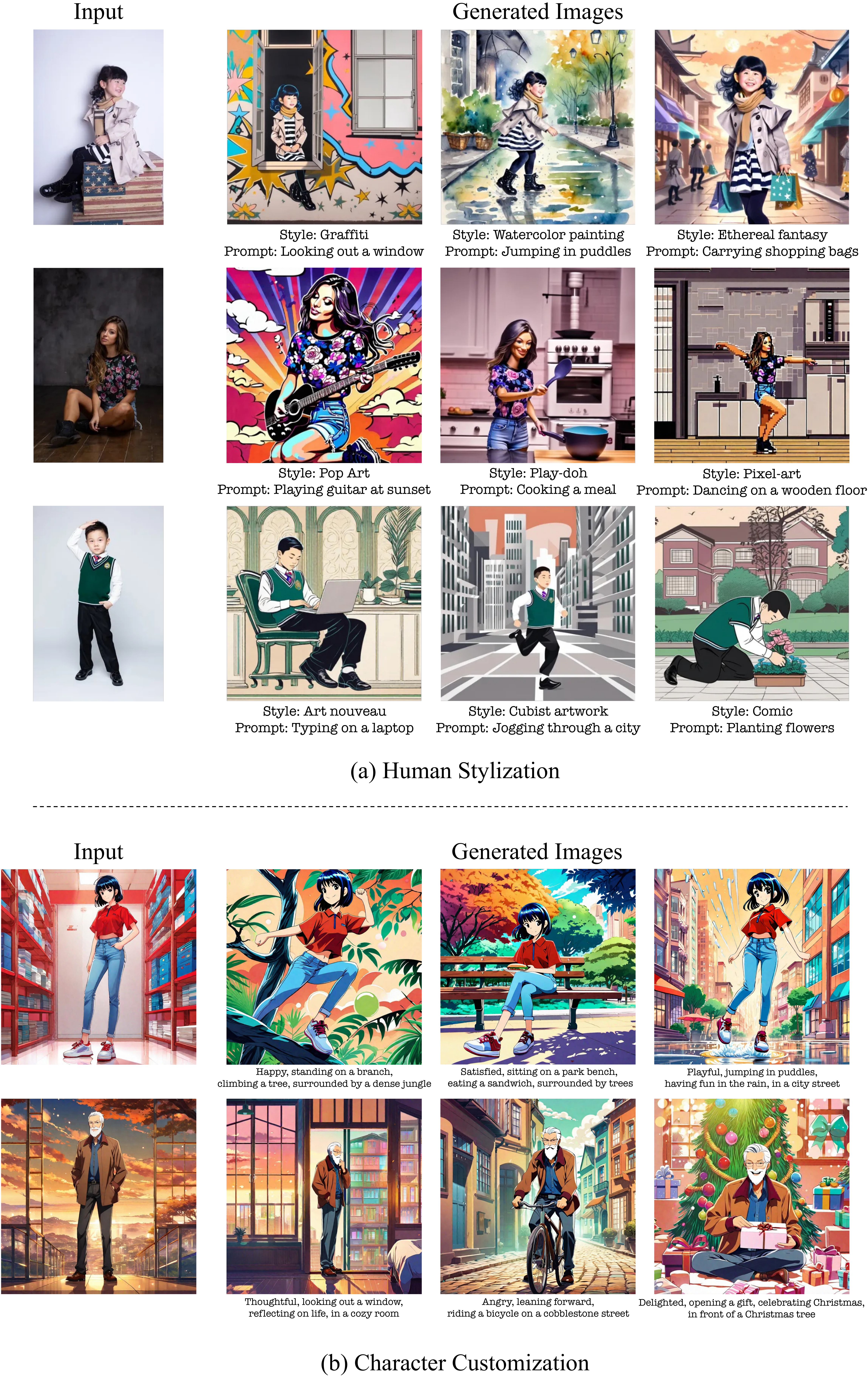} 
    \end{center}
    \vspace{-10pt}
    \caption{\textbf{Qualitative Results of Visual Persona for Human Stylization and Character Customization:} (a) Visual Persona can adapt to various stylization prompts while preserving the input's full-body identity. (b) Although Visual Persona is not trained for the character domain, our method can generalize to the character domain.}

    \label{qual:appendix_app_qual_2}
\end{figure*}

\begin{figure*}[t]
    \begin{center}
\includegraphics[width=0.9\linewidth]{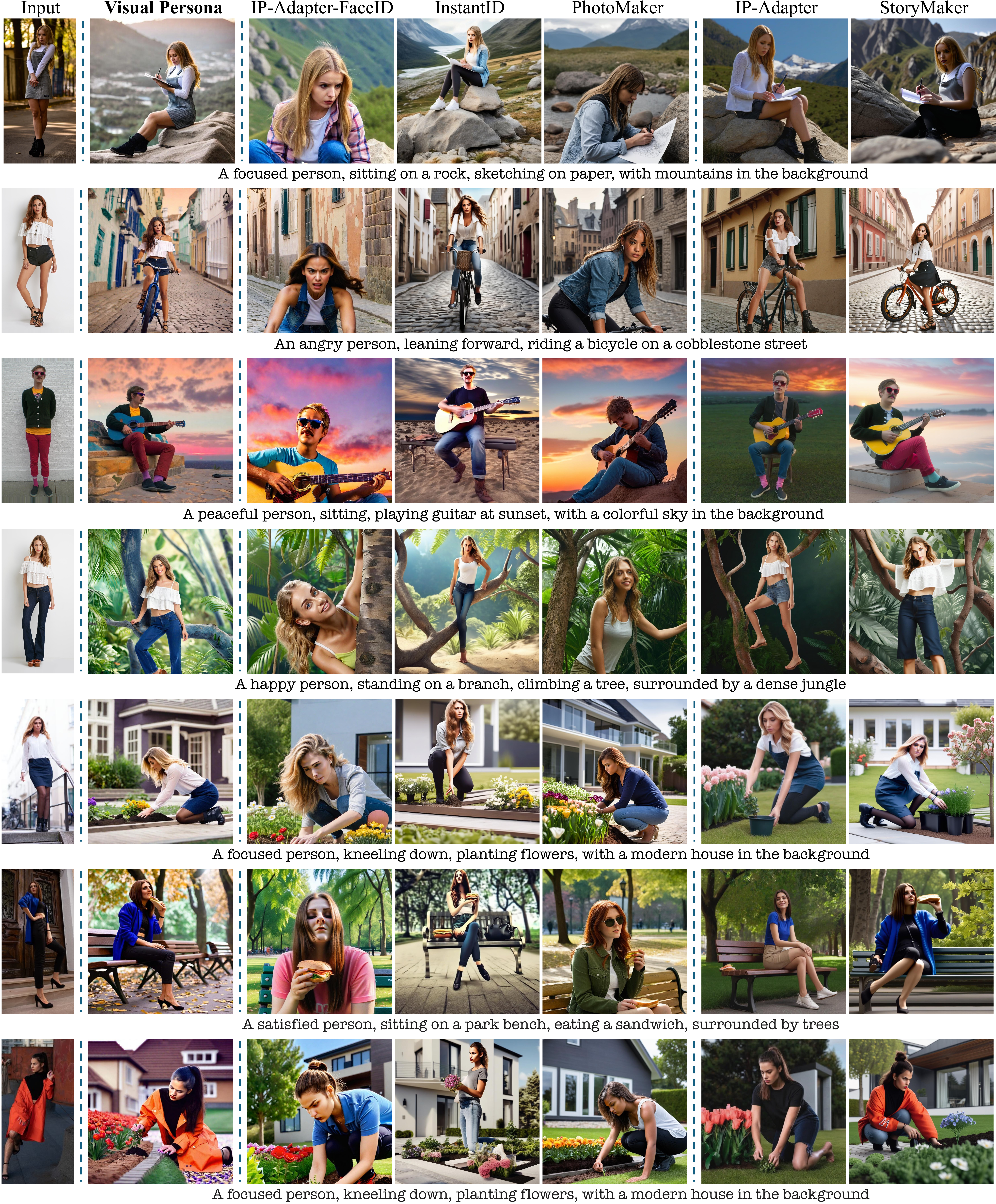} 
    \end{center}
    \vspace{-15pt}
    \caption{\textbf{Qualitative Comparison on SSHQ~\cite{fu2022stylegan} and PPR10K~\cite{liang2021ppr10k}:} We compare Visual Persona with IP-Adapter-FaceID~\cite{ye2023ip}, InstantID~\cite{wang2024instantid}, PhotoMaker~\cite{li2024photomaker}, IP-Adapter~\cite{ye2023ip}, and StoryMaker~\cite{zhou2024storymaker}.}

    \label{qual:appendix_qual_3}
\end{figure*}

More qualitative results of Visual Persona on SSHQ~\cite{fu2022stylegan} and PPR10K~\cite{liang2021ppr10k} are provided in Figure~\ref{qual:appendix_qual_1} and Figure~\ref{qual:appendix_qual_2}. Additional qualitative results on applications, including text-guided virtual try-on, human stylization, and character customization, are provided in Figure~\ref{qual:appendix_app_qual_1} and Figure~\ref{qual:appendix_app_qual_2}. Additional qualitative comparison results with~\cite{ye2023ip, wang2024instantid, li2024photomaker, zhou2024storymaker} are presented in Figure~\ref{qual:appendix_qual_3}.

\clearpage

\clearpage

\clearpage

\begin{figure}[t]
    \centering
    \includegraphics[width=0.8\linewidth]{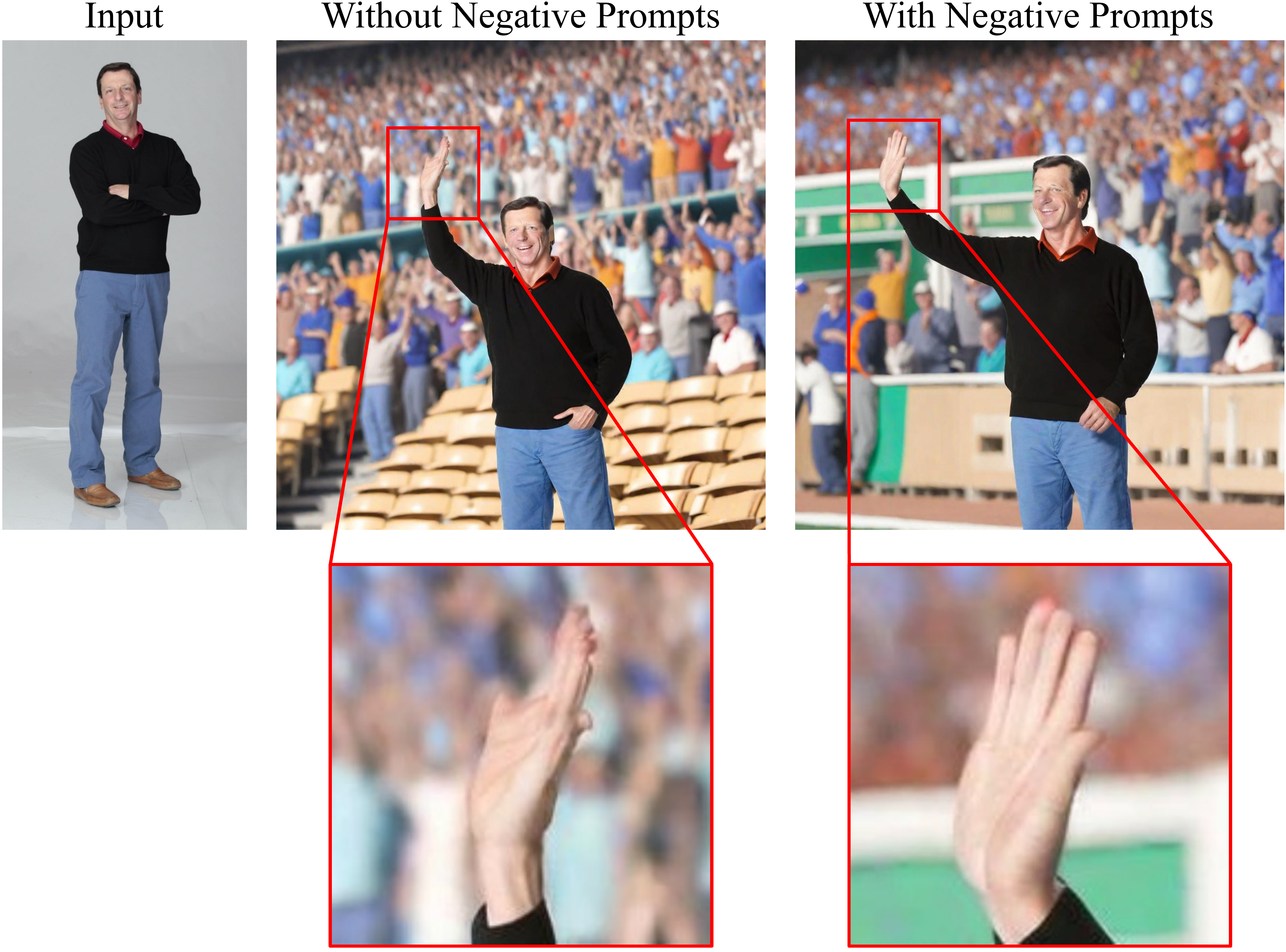} 
    \caption{\textbf{Limitation: Inaccurate Body Proportions.} The inherent challenge of generating bad body proportions in pre-trained T2I diffusion models can be alleviated through negative prompting.}
    \vspace{-5pt}
    
    \label{qual:limitation}
\end{figure} 
\begin{figure}[t]
    \centering
    \includegraphics[width=1.0\linewidth]{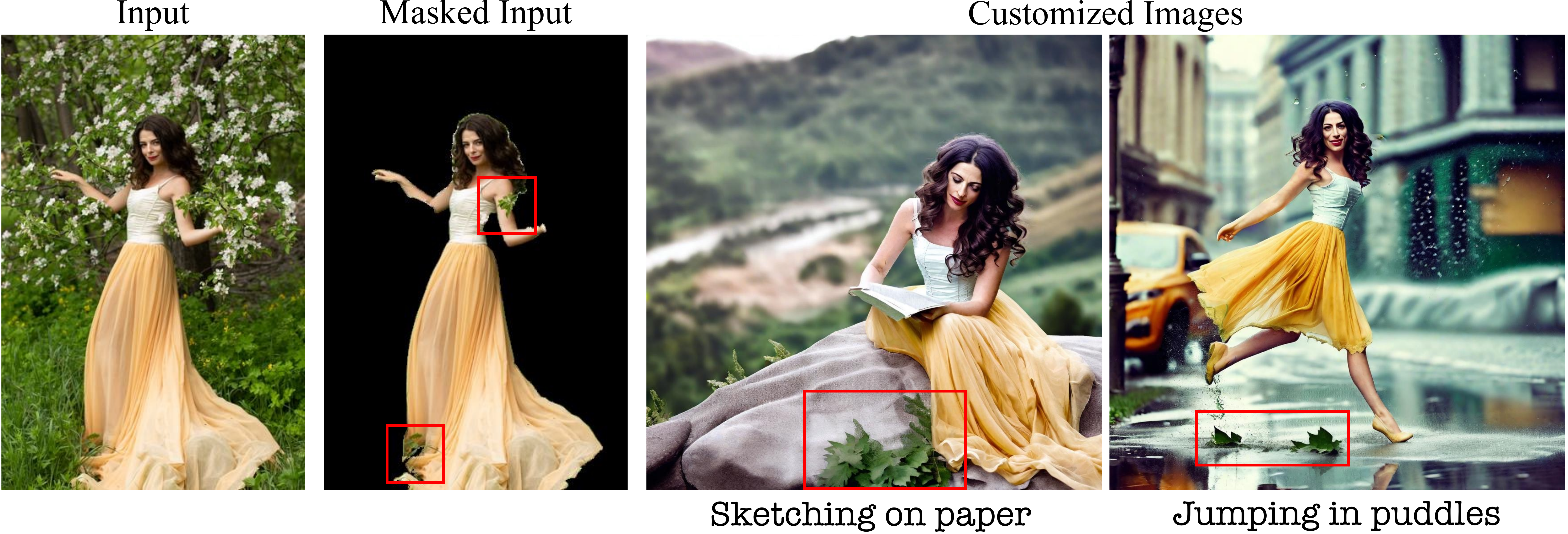} 
    \caption{\textbf{Limitation: Identity-Unrelated Attribute Leakage from Input.} When the input human is occluded by identity-unrelated elements, these elements are often included in the customized images.}
    \vspace{-5pt}
    
    \label{qual:limitation2}
\end{figure}

\section{Limitation}
\label{supp:limitation}
\paragrapht{Inaccurate Body Proportions.} SDXL~\cite{podell2023sdxl} inherently struggles to generate human images with accurate body proportions, often resulting in artifacts such as fused fingers or extra arms and legs. Figure~\ref{qual:limitation} provides an example of fused fingers. Since we leverage the pre-trained SDXL to maximize its generative capabilities, our model also inherits these issues. To alleviate this, we incorporate a negative prompt, including terms such as \textit{“disfigured, deformed, three arms, three legs, fused fingers, cloned face, bad proportions, bad anatomy.”} This negative prompt guides the pre-trained T2I diffusion model away from generating such undesired features through classifier-free guidance~\cite{ho2022classifier}. Figure~\ref{qual:limitation} shows that this negative prompt effectively enables the model to generate more natural and anatomically accurate human body proportions, without the need for additional training.

\paragrapht{Identity-Unrelated Attribute Leakage from Input.} Figure~\ref{qual:limitation2} shows that when the input human is occluded by identity-unrelated elements (e.g., background leaves or grass), the customized image includes these elements instead of filtering them out.In future work, we plan to address this by refining the foreground mask using body parsing models~\cite{li2020self, khirodkar2024sapiens}, which separate each part of the human body individually and more accurately isolate only the foreground, in contrast to the human matting method~\cite{huynh2024maggie}, which directly detects the whole human body.

\end{document}